\documentclass[nohyperref]{article}

\usepackage{microtype}
\usepackage{graphicx}
\usepackage{subfigure}
\usepackage{booktabs} 
\usepackage{hyperref}

\usepackage[accepted]{icml2022}

\usepackage{amsmath}
\usepackage{amssymb}
\usepackage{mathtools}
\usepackage{amsthm}
\usepackage[capitalize,noabbrev]{cleveref}
\usepackage[utf8]{inputenc} 
\usepackage[T1]{fontenc}   
\usepackage{url}            
\usepackage{booktabs}       
\usepackage{amsfonts}       
\usepackage{nicefrac}       
\usepackage{microtype}      
\usepackage{amsmath, amssymb, amsthm}
\usepackage{algorithm}
\usepackage{xcolor}
\usepackage{makecell}
\usepackage{rotating}

\usepackage{multirow}
\usepackage{overpic}
\usepackage{dsfont}
\usepackage{algorithmic}
\usepackage{graphicx}
\usepackage{pdflscape}
\usepackage{enumitem}
\setlist[enumerate]{leftmargin=*}
\setlist[itemize]{leftmargin=*}

\def\argmin{\mathop{\textnormal{argmin}}}
\def\argmax{\mathop{\textnormal{argmax}}}

\newcommand{\E}{\mathop{\mathbb{E}}}
\newcommand{\R}{\mathbb{R}}

\def\Z{\mathcal{Z}}
\def\F{\mathcal{F}}

\newtheorem{theorem}{Theorem}
\newtheorem{assumption}{Assumption}

\newtheorem{lemma}{Lemma}

\newtheorem{corollary}{Corollary}

\theoremstyle{remark}
\newtheorem{remark}{Remark}

\newcommand{\vloss}[2]{v\left(#1; #2\right)}

\newcommand{\grp}[1][i]{D_{#1}}
\newcommand{\grpemp}[1][i]{\hat D_{#1}}

\newcommand{\mix}{\mathbf{q}}
\newcommand{\mixt}[1][t]{\mathbf{q}_{#1}}
\newcommand{\tht}{\theta}
\newcommand{\thtt}[1][t]{\theta_{#1}}
\newcommand{\thtavg}[1][T]{\bar \theta_{#1}}
\newcommand{\grplosses}[1][t]{\mathbf{u}_{#1}}
\newcommand{\ent}[1]{\textsc{Ent}(#1)}
\newcommand{\kl}[2]{\textsc{KL}(#1||#2)}
\newcommand{\rad}{\mathcal{R}}

\theoremstyle{definition}

\usepackage[textsize=tiny]{todonotes}

\icmltitlerunning{Active Sampling for Min-Max Fairness}

\begin{document}

\twocolumn[
\icmltitle{Active Sampling for Min-Max Fairness}

\begin{icmlauthorlist}
\icmlauthor{Jacob Abernethy}{gatech}
\icmlauthor{Pranjal Awasthi}{google}
\icmlauthor{Matth{\"a}us Kleindessner}{amazon}
\icmlauthor{Jamie Morgenstern}{uw}
\icmlauthor{Chris Russell}{amazon}
\icmlauthor{Jie Zhang}{uw}
\end{icmlauthorlist}

\icmlaffiliation{gatech}{Georgia Tech, USA}
\icmlaffiliation{google}{Google, USA}
\icmlaffiliation{amazon}{Amazon Web Services, Germany}
\icmlaffiliation{uw}{University of Washington, USA}

\icmlcorrespondingauthor{J. Abernethy}{prof@gatech.edu}
\icmlcorrespondingauthor{P. Awasthi}{pranjalawasthi@google.com}
\icmlcorrespondingauthor{M. Kleindessner}{matkle@amazon.de}
\icmlcorrespondingauthor{J. Morgenstern}{jamiemmt@cs.washington.edu}
\icmlcorrespondingauthor{C. Russell}{cmruss@amazon.de}
\icmlcorrespondingauthor{J. Zhang}{claizhan@uw.edu}

\icmlkeywords{Machine Learning, Min-Max Fairness}

\vskip 0.3in
]

\printAffiliationsAndNotice{}  %

\begin{abstract}

We propose 
simple 
active
sampling and reweighting strategies for optimizing min-max fairness that can be applied to 
any classification or regression model learned via loss minimization. 
The key intuition behind our 
approach is to use at each timestep a datapoint from the group that is  worst off under the current model 
for updating 
the model. 
The ease of implementation and the generality of our robust formulation make it an attractive option for improving model performance 
on disadvantaged groups. 
For convex learning problems, such as linear or logistic regression, we provide a fine-grained analysis, 
proving the rate of convergence to a min-max fair solution.
\end{abstract}

\section{Introduction}\label{section_intro}

A model trained on a dataset 
containing multiple demographic groups
typically has unequal error rates across the different groups, either because some 
groups 
are underrepresented in the training data or  the underlying learning task is inherently harder for particular groups. Many existing fairness notions aim to equalize the performance on different demographic groups~\citep[e.g., ][]{hardt2016equality,zafar2019JMLR}, which can  result in 
deliberately down-grading the performance 
on the better off groups 
and unnecessarily reducing overall performance.
 This degradation of performance  %
 can correspond to a loss of access to relevant services, which is  referred to as ``leveling-down'' in law and ethics where it has received substantial criticism~\citep{holtug1998egalitarianism,temkin2000equality,doran2001reconsidering,mason2001egalitarianism,
brown2003giving,christiano2008inequality}. In contrast, min-max notions of fairness offer an alternative approach. These notions ``level-up'', by adopting an optimization perspective that prioritizes improving the model's performance on the group for which performance is 
the worst. Such optimizations only degrade the performance on a group if it will improve the performance on  the 
worst off 
group~\citep{martinez2020, diana2021}.

In this paper, 
we propose simple and theoretically principled algorithms for  min-max fairness (formally defined in Section~\ref{section_setup}). We provide a general template in Algorithm~\ref{alg:basic_outline}. The key idea underlying our approach is to adaptively sample or reweight data from 
worst off 
groups. The intent is that by providing additional data for 
these 
groups, or increasing the weights associated with them, we improve the model's performance on these groups. This is a compellingly simple approach to mitigating loss disparity between groups, and  we show that for convex learning problems such methods indeed provably minimize the maximum per-group loss.

We consider two concrete variants of Algorithm~\ref{alg:basic_outline}. The first one~(Algorithm~\ref{minimax_sgd}) simply samples a point from the 
worst off 
group and updates model parameters via stochastic gradient descent. We show in Theorem~\ref{thm:sgd_basic} that 
Algorithm~\ref{minimax_sgd} 
converges to a min-max fair solution at a rate of $\sim 1/\sqrt{T}$~(as a function of the number of iterations~$T$), assuming a  convex loss function. Our %
second 
approach~(Algorithm~\ref{minimax_sgd-optim}), {\em accelerated min-max gradient descent},   converges to 
a 
min-max fair solution at a faster rate of $\sim 1/T$. 

The accelerated 
Algorithm~\ref{minimax_sgd-optim} is based on an adaptive reweighting scheme and performs in each iteration a gradient descent step for optimizing a weighted population loss. 
In contrast, 
Algorithm~\ref{minimax_sgd} samples, in each iteration, a  datapoint from the 
worst off 
group and uses it for performing a stochastic gradient descent step. 
While  Algorithm~\ref{minimax_sgd-optim} achieves a faster rate of convergence, Algorithm~\ref{minimax_sgd} is
conceptually simpler, 
easier to implement and efficient in practice due to the stochastic nature of the updates. We 
also provide finite sample generalization bounds for 
Algorithm~\ref{minimax_sgd}. 
We present an 
empirical 
evaluation of both our proposed 
algorithms  
in Section~\ref{section_experiments}.

\newcommand{\SpaceInAlg}{1pt}
\begin{algorithm}[t!]
  \caption{A generic group-specific loss aware sampling strategy}
  \label{alg:basic_outline}
\begin{algorithmic}[1]
\vspace{\SpaceInAlg}
\STATE Initialize 
classifier / regressor 
$f$ with  initial model %
\vspace{\SpaceInAlg}
\STATE \textbf{repeat:}
\vspace{\SpaceInAlg}
\STATE \hspace{4mm} Determine the group for which the current model $f$ has the highest loss
\vspace{\SpaceInAlg}
\STATE \hspace{4mm} Sample a labelled datapoint 
 from that group and use it to update $f$
\vspace{\SpaceInAlg}
\STATE
(optional) Set $f$ to the  average over all past $f$ 
\STATE Return $f$
\end{algorithmic}
\end{algorithm}

\section{Preliminaries}\label{section_setup}

Let $\Z$ be our data space; any $z \in \Z$ represents a population sample containing both the observed features and the unobserved label. We assume that there are $g$ disjoint demographic groups, indexed by the set $[g]:=\{1,\ldots,g\}$.\footnote{
Sometimes we want to be fair w.r.t. demographic groups that are not disjoint, e.g., to men and women and also to old and young people. In this case, as it is standard in the literature, we simply consider all intersections of these groups, i.e., young females, young males, etc..
}

Let $\grp[1], \ldots, \grp[g]$ 
be 
a family of distributions over $\Z$, where $\grp[j]\in\Delta(\Z):=\{\text{distributions } \text{over $Z$}\}$ is the distribution of the data for group $j$. Let $\mix \in \Delta_g:=\{\mathbf{v}\in[0,1]^g: \sum_{i=1}^g\mathbf{v}_i=1\}$ denote a vector of mixture weights over groups. Given any $\mix$, we 
define the mixture distribution $\grp[\mix]$ on $\Z$ as follows: we first sample a group index $i \sim \mix$, and then we sample $z \sim \grp[i]$ from the randomly chosen group $i$.

The models considered are based on a parameterized family of functions $\F := \{ f_\tht : \tht \in \Theta \}$, where each $f_\theta$ operates on examples $z \in \Z$ and $\Theta \subset \R^d$ is a $d$-dimensional parameter space. We assume that $\Theta$ is a compact convex set. We evaluate each $f_\theta$ according to a loss function $\ell : \F \times \Z \to \R$, and we assume  the loss~$\ell(f_\theta; z)$ to be convex in $\tht$ for any fixed $z \in \Z$: 

\begin{assumption}\label{assumption_convexity} 
For any $z\in   \mathcal{Z}$, the function $\theta \mapsto \ell(f_{\theta};z)$ is convex in $\theta$. 
\end{assumption}

Assumption~\ref{assumption_convexity} is satisfied in several common scenarios, e.g., when $f_\theta$ 
is linear in
$\theta$ and $\ell$ is the standard logistic loss or hinge loss (binary classification),  
the cross entropy loss composed with softmax activation (multiclass classification), 
or
the squared loss (regression). While we use $\nabla \ell$ throughout the paper to refer to the gradient, we do not strictly need %
$\theta \mapsto \ell(f_{\theta};z)$ 
to be differentiable as we may consider any sub-gradient instead.

Given a distribution $\grp[] \in \Delta(\Z)$ and any $\tht \in \Theta$, we define the expected loss of $\tht$ with respect to $\grp[]$ as
\[
  \vloss{\tht}{\grp[]} := \E_{z \sim \grp[]} \ell(f_\tht, z).
\]
Similarly, given mixture 
weights 
$\mix \in \Delta_g$, we consider the performance of $f_\tht$ with respect to $\grp[\mix]$. Thus
\[
     \vloss{\theta}{\grp[\mix]} =  \E_{i \sim \mix} \left[ \E_{z \sim \grp} [ \ell( \theta, z) ] \right ]
    = \sum_{i=1}^g \mix(i) \E_{z \sim \grp[i]} \ell(\theta, z).
\]
For each group $i \in [g]$ we assume we have an IID sample of $m_i$ examples $z_1, \ldots, z_{m_i} \sim D_i$. We use $\grpemp[i]$ to represent the empirical distribution over these samples. Hence, 
\[
  \vloss \tht {\grpemp[i]} = \frac 1 {m_i} \sum_{j=1}^{m_i} \ell(f_\tht, z_j).
\]
\def\proj{\textsc{Proj}}
Throughout, we use the notation $\proj_K(x)$ to refer to the $\ell_2$-projection of point $x \in \R^d$ onto compact convex set~$K \subset \R^d$, that is  $\proj_K(x) := \argmin_{y \in K} \|x - y\|_2$.

Our goal is to learn 
a model $f_{\theta^\star}$ 
that is min-max fair w.r.t. the $g$ demographic groups.  
This means that $\theta^\star$ satisfies 
\begin{align*}
  \max_{i \in [g]} \vloss{{\theta^\star}}{\grp} =  \inf_{\theta \in \Theta} \max_{i \in [g]} \vloss{{\theta}}{\grp}.
\end{align*}

\begin{remark}\label{remark_mixture_distri}
A 
criticism of the min-max approach to fairness is that it puts too much focus on improving the performance of a single group. If some class $j$ is particularly hard to learn, so that any $\theta \in \Theta$ will have a large value $ \vloss{{\theta}}{\grp[j]}$, larger than for all other classes~$j'$, then 
the training procedure will aim to optimize the loss with respect to only $\grp[j]$. This is a reasonable concern, but we can mitigate it by considering  blended distributions: let $p \in [0,1]$ be a trade-off parameter, and define the mixture distribution
\[
    \tilde{\grp}^p := (1-p) \grp + p \grp[\hat \mix],
\]
where $\grp[\hat \mix]$ is the true population-level distribution; 
that is, 
the full population is made up of a mixture of subpopulations $\grp[1], \ldots, \grp[g]$ weighted by the true mixture $\hat \mix$. If we run our min-max fairness procedures on the blended distributions~$\tilde{\grp}^p$ instead of on $\grp$, we are biasing our algorithm, with bias parameter~$p$, to focus more on the full population and less on any particular group. %
\end{remark}

\section{Algorithms and Analysis}\label{section_analysis}

Here we present formal versions of Algorithm~\ref{alg:basic_outline}, and analyze their theoretical performance. To formalize Algorithm~\ref{alg:basic_outline} fully, we describe \emph{how} we evaluate which group has the highest loss in each iteration, and how we either sample  additional data from that group or reweight and update the model.
Both algorithms use their updating schemes to reduce their \emph{training} error on the min-max objective, and while we also present theorems which bound the test error as well, this work does not use reweighting or resampling to explicitly decrease generalization error.

\subsection{Stochastic Optimization}

Algorithm~\ref{minimax_sgd} maintains a validation set, 
that is 
a fixed comparison sample set on which the group-specific loss is repeatedly measured. It samples a fresh point from the group with highest loss on the comparison set, then takes a single gradient step in the direction of that fresh sample. Its performance is governed by two quantities: the regret term, which decreases with the number of iterations $T$, and the uniform deviation bound of the comparisons of group-specific loss.

\def\update{\textsc{Update}}

\begin{algorithm}[h]
   \caption{Min-max Stochastic Gradient Descent}
   \label{minimax_sgd}
\begin{algorithmic}[1]
\vspace{1mm}
 \STATE {\bfseries Init:} $\thtt[1]\in \Theta$ arbitrary
 \FOR{$t=1 \ldots T-1$}

 \STATE  compute $\displaystyle i_t=\argmax_{i\in[g]} \vloss{{\thtt[t]}}{\grpemp[i]}$
  \STATE sample $z_t \sim D_{i_t}$
  \STATE compute $\nabla_t \leftarrow \nabla_{\tht} \ell(f_{\thtt}; z_t)$
 \STATE update $\thtt[t+1] \leftarrow \proj_\Theta(\thtt - \eta \nabla_t)$
\ENDFOR
\STATE return $\thtavg = \frac{\sum_{t=1}^T  \theta_t}{T}$
 
\end{algorithmic}
\end{algorithm}

As is common in gradient descent, the following proof assumes that the  Lipschitz constant $L$ and a domain radius~$W$ are known. When this is not the case, the step size $\nu$ is typically tuned empirically.

\begin{theorem}\label{thm:sgd_basic}
Assume we have a function $\rad_\delta = \rad(m_1, \ldots, m_g; \delta)$ which guarantees that $$\sup_{\theta \in \Theta}\max_{i \in [g]} | \vloss \tht \grp - \vloss \tht \grpemp | \leq \rad_\delta$$ with probability at least $1-\delta$. 
Let $W:= \sup_{\theta \in \Theta} \|\theta - \theta_1\|_2$ and $L := \sup_{\theta \in \Theta}\max_{i \in [g]} \|\nabla_\theta \vloss{\theta}{\grp[i]}\|_2$.
With 
$\eta:=\textstyle \frac{W}{L \sqrt T}$,  
Algorithm~\ref{minimax_sgd} ensures that
$$
    \E_{z_{1:T}}\left[\max_{i \in [g]} \vloss{{\thtavg}}{\grp}\right] \leq \inf_{\theta \in \Theta} \max_{i \in [g]} \vloss{{\theta}}{\grp} +  \frac{WL}{\sqrt{T}} + 2\rad_{T\delta}
$$
with probability at least $1-\delta$.
\end{theorem}

\begin{figure*}[t]

  \caption{Main steps in the proof of Theorem~\ref{thm:sgd_basic}.\label{figure_proof_Theorem1} The proof proceeds along the lines of the classical online-to-batch conversion \citep{cesa2004generalization}, 
  but hinges on a few additional tricks.}
  \begin{eqnarray*}
   & &  \textstyle 
   \E_{z_{1:T}}\left[ \max_{i \in [g]} \vloss{{\thtavg}}{\grp[i]} \right]\\
    \left(\text{\small Jensen's inequality}\right) \quad \quad   
    & \leq  & \textstyle
      \E_{z_{1:T}}\left[ \frac 1 T \max_{i \in [g]}  \sum_{t=1}^T  \vloss{{\thtt[t]}}{\grp[i]} \right] \\
      \left(\text{\small max sum $\leq$ sum max}\right) \quad \quad    
      & \leq & \textstyle
      \E_{z_{1:T}}\left[  \frac 1 T  \sum_{t=1}^T  \left( \max_{i \in [g]} \vloss{{\thtt[t]}}{\grp[i]} \right) \right] \\
      \left(\text{\small deviation between $\grp[]$, $\grpemp[]$ + union bound}\right) \quad \quad    
      & \leq & \textstyle
        \E_{z_{1:T}}\left[  \frac 1 T  \sum_{t=1}^T  \max_{i \in [g]} \vloss{{\thtt[t]}}{\grpemp[i]} \right] + \rad_{T\delta} \\
      \left(\text{\small definition of $i_t$}\right) \quad \quad  
    & = & \textstyle
      \E_{z_{1:T}}\left[ \frac 1 T   \sum_{t=1}^T  \vloss{{\thtt[t]}}{\grpemp[i_t]} \right] + \rad_{T\delta}\\
      \left(\text{\small additional deviation between $\grpemp[]$, $\grp[]$}\right) \quad \quad  
    & \leq & \textstyle
      \E_{z_{1:T}}\left[ \frac 1 T   \sum_{t=1}^T  \vloss{{\thtt[t]}}{\grp[i_t]} \right] + 2\rad_{T\delta}\\
      \left(\text{\small since $z_t \sim \grp[i_t]$ + outer expectation }\right) \quad \quad  
    & = & \textstyle
      \E_{z_{1:T}}\left[  \frac 1 T  \sum_{t=1}^T  \ell(f_{\thtt};z_t) \right] + 2\rad_{T\delta}\\
      \left(\text{\small apply OGD regret bound}\right) \quad \quad  
    & \leq &\textstyle
      \E_{z_{1:T}}\left[  \frac 1 T  \sum_{t=1}^T  \ell(f_{\theta^\star};z_t) + \frac{\eta L^2}{2} + \frac {W^2} {2T\eta} \right] + 2\rad_{T\delta}\\
      \left(\text{$\eta:=\textstyle \frac{W}{L \sqrt T}$, $S^T := \{z_1, \ldots, z_T\}$ }\right) \quad \quad  
    & = &\textstyle
      \E_{z_{1:T}}\left[ \vloss{{\theta^\star}}{S^{T}} \right] +  \frac{WL}{\sqrt T} + 2\rad_{T\delta}\\
    & = &\textstyle
       \vloss{{\theta^\star}}{\grp[\tilde \mix]} + \frac{WL}{\sqrt T} + 2\rad_{T\delta}\\
    & \leq & \textstyle \max_{i \in [g]}  \vloss{{\theta^\star}}{D_i} + \frac{WL}{\sqrt T} + 2\rad_{T\delta} 
  \end{eqnarray*}
 \end{figure*}

\begin{proof}
This bound is obtained by combining a number of classical results from empirical process theory, as well as common tricks from using online convex optimization tools to solve min-max problems. This sequence of steps is given in Figure~\ref{figure_proof_Theorem1}, with further discussion here.

One observes that we need to swap between $\vloss{{\thtt[]}}{\grpemp[]}$ and $\vloss{{\thtt[]}}{\grp[]}$ on two separate inequalities, and on each we have to add the deviation bound $\rad_{T\delta}$; the $T$ factor is necessary because we need a union bound over all $T$ rounds. We then replace $\vloss{{\thtt[t]}}{\grp[i_t]}$ with $\ell(f_{\thtt};z_t)$, which is valid since $\theta_t$ is independent of $z_t$, $z_t$ is distributed according to $\grp[i_t]$, and we have the outer expectation over all $z_1, \ldots, z_T$ (more details on this technique can be found in the paper of \citealp{cesa2004generalization}). Next, since the $\theta_t$'s are chosen using the Online Gradient Descent (OGD) algorithm on loss functions $h_t(\cdot) := \ell(f_{\cdot}; z_t)$, we can immediately apply the OGD regret bound---see \citet{hazan2019introduction} for details.

  The most subtle part of this proof may be the final two observations. The sequence $z_{1:T}$ is generated stochastically and sequentially, where each $z_t$ may depend on the previous samples chosen. But in the end, the sample $S^{T}$ is produced by taking some combination of samples from the various $\grp[1], \ldots, \grp[g]$, and ultimately we \emph{marginalize} the quantity $\vloss{{\theta^\star}}{S^{T}}$ over the randomness generated by $z_1, \ldots, z_T$. On average, the $z$'s in $S^T$ will have been drawn from some mixture over the various groups, and we refer to 
  those 
  mixture weights 
  as 
  $\tilde \mix$. It then follows by this observation that $\E_{z_{1:T}}\left[ \vloss{{\theta^\star}}{S^{T}} \right] = \vloss{{\theta^\star}}{\grp[\tilde \mix]}$. Finally, since $\vloss{{\theta^\star}}{\grp[\tilde \mix]} = \E_{i \sim \tilde \mix} \vloss{{\theta^\star}}{\grp[i]}$, we  upper bound $\E_{i \sim \tilde \mix}$ with $\max_{i}$ to complete the proof. 
\end{proof}

While 
not the focus of our paper, it is easy enough to give a uniform deviation bound as Theorem~\ref{thm:sgd_basic} employs.
 
\begin{lemma}
  With probability at least $1-\delta$,  for every $\theta \in \Theta$ and for every $\mix \in \Delta_g$ it holds that
  \[
  \vloss{\theta}{ \grpemp[\mix]} \leq \vloss{\theta}{\grp[\mix]} + c\sqrt{\frac{\text{P-dim}(\Theta) \log (g \cdot m_{\min}/\delta)}{\min_i m_i}}
  \]
 where $c > 0$ is some constant and $\text{P-dim}$ is the pseudo-dimension of the class~\citep{pollard1990empirical}.
\end{lemma}
\begin{proof}
This follows from a standard uniform convergence argument over $\Theta$ for any fixed group $i$, as $\hat D_{i}$ is a sample of $m_i$ IID points drawn from $D_i$. Taking a union bound over all $g$ groups yields the 
bound.
\end{proof}
This implies that the total error of using Algorithm~\ref{minimax_sgd} is comprised of two terms, one upper bounding the optimization error (which decays with $1/\sqrt{T}$), and the generalization error, which is governed by the sample size of the smallest dataset across all groups.

A key benefit of Theorem~\ref{thm:sgd_basic} is that it provides both an \emph{optimization} guarantee as well as a \emph{sample complexity} bound. The proof's core is the classical ``online to batch conversion'' \citep{cesa2004generalization} that provides generalization based on regret bounds, combined with tools from min-max optimization. 

One downside of this method is that it relies on having two sources of data: a ``comparison set'' $\grpemp$ for each $i \in [g]$, as well as the ability to draw fresh (independent) samples from each $\grp$. Alternatively, we consider a version that only focuses on training error that reweights rather than samples fresh data, by considering $z_t$  drawn from $\grpemp$. This variant will still allow for the min-max empirical risk to decay at the standard 
$1/\sqrt{T}$
rate.

\begin{corollary}
Consider a version of Algorithm~\ref{minimax_sgd} that, on line~4, draws samples IID from the empirical distribution~$\grpemp[i_t]$ as opposed to fresh samples from $\grp[i_t]$. Then, with $W,L$ defined as in Theorem~\ref{thm:sgd_basic}, we have 
$$
    \E_{z_{1:T}}\left[\max_{i \in [g]} \vloss{{\thtavg}}{\grpemp}\right] \leq \inf_{\theta \in \Theta} \max_{i \in [g]} \vloss{{\theta}}{\grpemp} +  \frac{WL}{\sqrt{T}}.
$$
\end{corollary}\label{cor:sgd_emperr}

\begin{remark}[Mini-batching\label{remark_minibatches}]
Many online training scenarios use mini-batch gradient updates, where instead of a single sample a set of samples is taken, an average gradient is computed across these samples, and 
the average gradient is  
used to update the current parameter estimate. Indeed, it requires a straightforward modification to implement mini-batch training in Algorithm~\ref{minimax_sgd}. While this may have practical benefits, providing faster empirical training times,  we 
note that this is not likely to provide improved theoretical guarantees. Our convergence guarantee in Theorem~\ref{thm:sgd_basic} still applies in the mini-batch setting, with convergence depending on the number of updates $T$, rather than the total amount of data used. Batches of size $k$ then require $k$ times more data overall for the same convergence guarantee. One might hope for a decrease in variance from the mini-batch averaging, and indeed this often empirically leads to better convergence, though not promised by our results.
\end{remark}

\subsection{Accelerated Optimization}
Algorithm~\ref{minimax_sgd-optim}'s 
optimization error shrinks much faster as a function of $T$. However,  it is non-stochastic and has a more complex update rule. %
This algorithm 
explicitly maintains a distribution over groups which it updates relative to the current group losses, increasing the probability mass assigned to groups with higher loss. Then the algorithm takes a gradient step with respect to the full, weighted distribution over (empirical) group distributions. While each iteration requires a full pass over the data, the convergence rate is $O(1/T)$ rather than $O(1/\sqrt T)$. 

 Unlike Algorithm~\ref{minimax_sgd}, 
 Algorithm~\ref{minimax_sgd-optim} 
 does not have a natural ``sampling'' analogue. The update rule is with respect to the entire weighted empirical distribution, rather than a single datapoint.
Below we present the main algorithmic guarantee associated with the algorithm. 
See Appendix~\ref{sec:accel} for the proof. 

\begin{algorithm}[h]
   \caption{Accelerated Min-max Gradient Descent}
   \label{minimax_sgd-optim}
\begin{algorithmic}[1]
\vspace{1mm}
 \STATE {\bfseries Init:} $\mixt[0] = (\frac1g, \ldots, \frac1g)$, $\thtt[1] \in \Theta$ arbitrary, $\nabla_{0} = \mathbf{0}$
 \FOR{$t=1 \ldots T$}
 \STATE Compute $\grplosses(i) \gets \vloss{{\thtt[t]}}{\grpemp[i]}$, for $i=1, \ldots, g$
 \STATE Update $\mixt[t](i) \gets \mixt[t-1](i)\exp(\gamma \grplosses(i))$, for $i=1, \ldots, g$
 \STATE Normalize $\mixt[t] \gets \frac{\mixt[t]}{\|\mixt[t]\|_1}$
 \STATE Compute $\nabla_{t} \gets \nabla_\theta \vloss{{\thtt[t]}}{ \grpemp[\mixt]}$
 \STATE Update $\thtt[t+1] \leftarrow \proj_\Theta(\thtt[t] - 2\eta \nabla_{t} + \eta \nabla_{t-1})$
\ENDFOR
\STATE return $\thtavg = \frac{\sum_{t=1}^T \theta_t}{T}$
 
\end{algorithmic}
\end{algorithm}

\begin{theorem}\label{theorem_advanced_alg}
Algorithm~\ref{minimax_sgd-optim}, with parameters $\eta = \frac{W}{L\sqrt{\log g}}$ and $\gamma = \frac{\sqrt{\log g}}{WL}$, outputs $\thtavg$ that satisfies
$$
        \max_{i \in [g]} \vloss{{\thtavg}}{\grpemp[i]} \leq \inf_{\theta \in \Theta} \max_{i \in [g]}\vloss{{\theta}}{\grpemp[i]} + \frac{2WL\sqrt{\log g} }{T},
$$
where $W$ and $L$ are defined as in Theorem~\ref{thm:sgd_basic}.

\end{theorem}\label{thm:minimax-sgd}

\begin{remark}[Averaging versus final iterate]
A careful reader may note that  Algorithms~\ref{minimax_sgd} and~\ref{minimax_sgd-optim}  output a time-weighted average $\thtavg$, whereas in typical online training methods one simply outputs the final iterate~$\theta_T$. Indeed, for the min-max framework we propose, our theory requires returning the average iterate. Some work exists on last-iterate convergence for special cases of min-max optimization \citep{pmlr-v132-abernethy21a}, but this is beyond the scope of the present work. 
\end{remark}

\section{Related Work}\label{section_related_work}

\newcommand{\scaleA}{0.21}
\newcommand{\heightA}{3.2cm}

\begin{figure*}[t]
    \centering
    \includegraphics[height=\heightA]{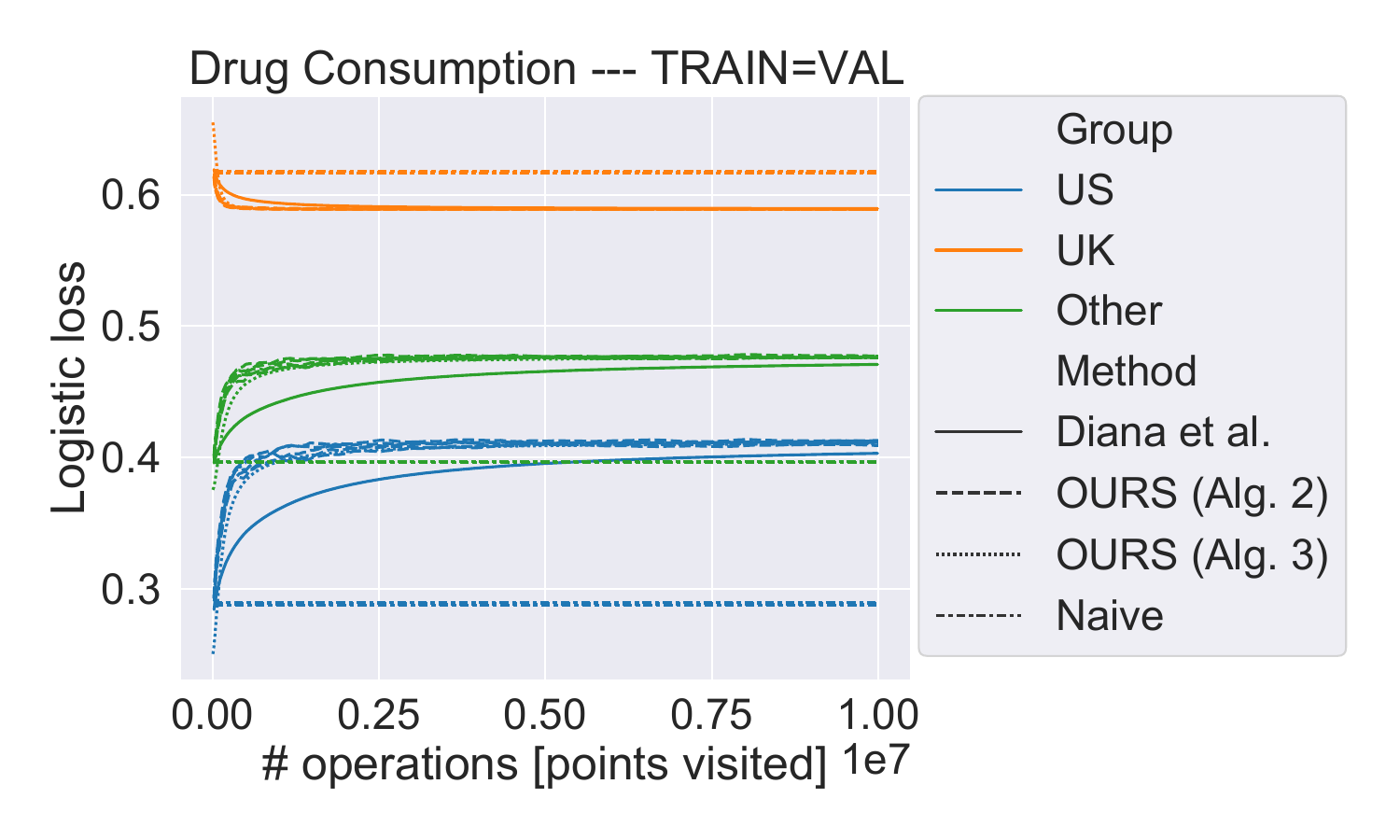}
    \hspace{6mm}
   \includegraphics[height=\heightA]{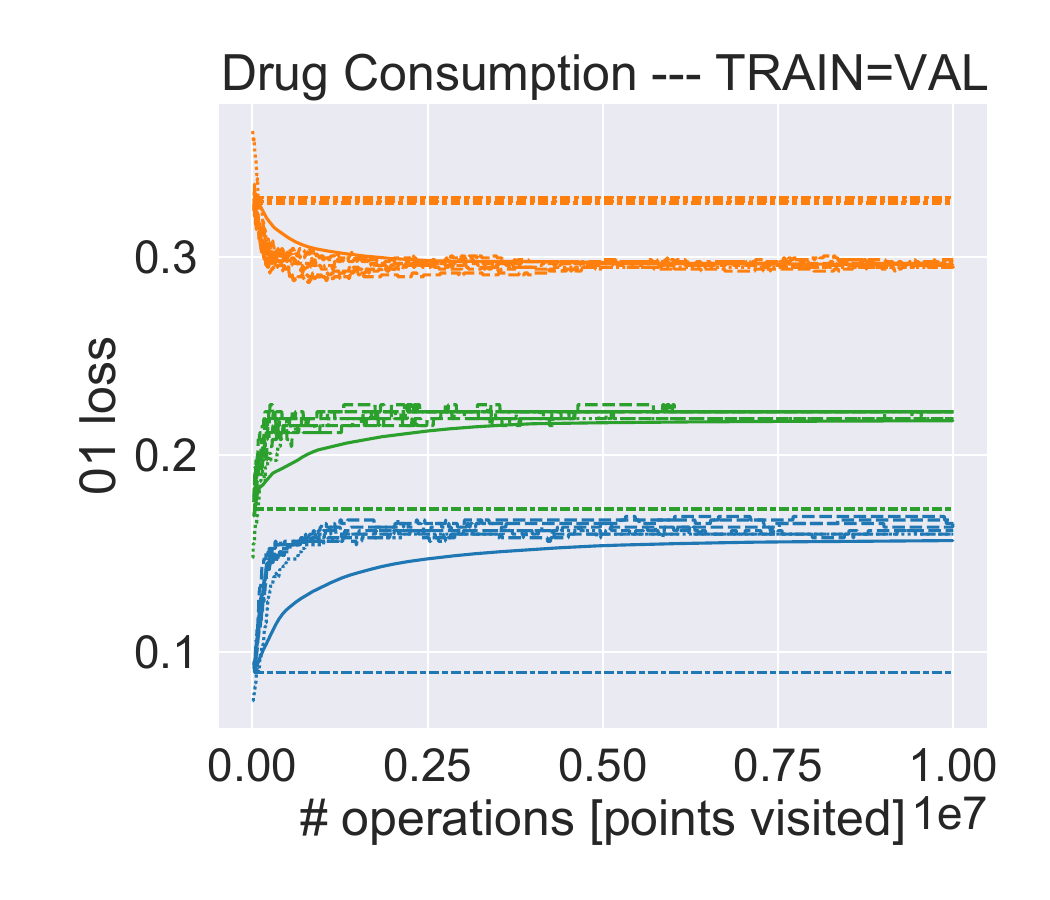}
    \hspace{6mm}
\includegraphics[height=\heightA]{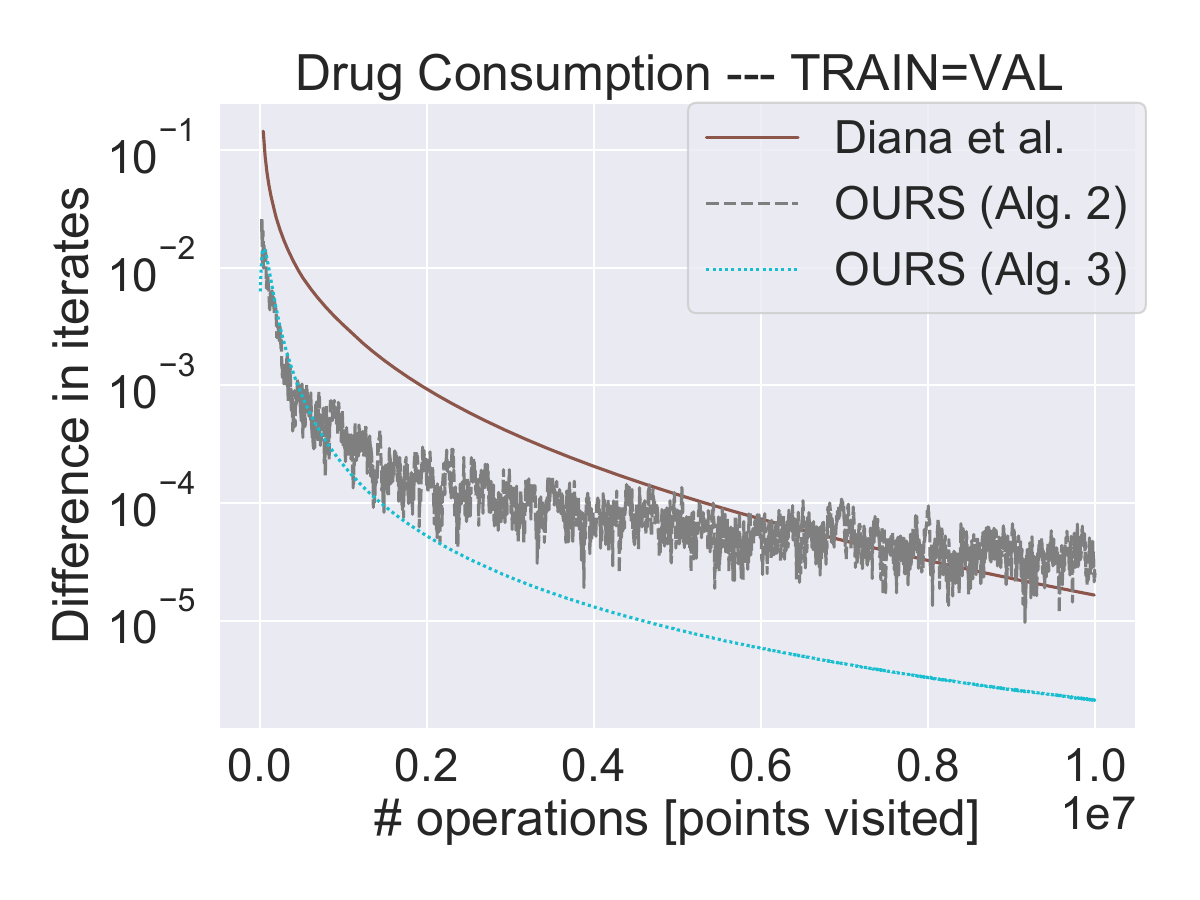}

    \includegraphics[height=\heightA]{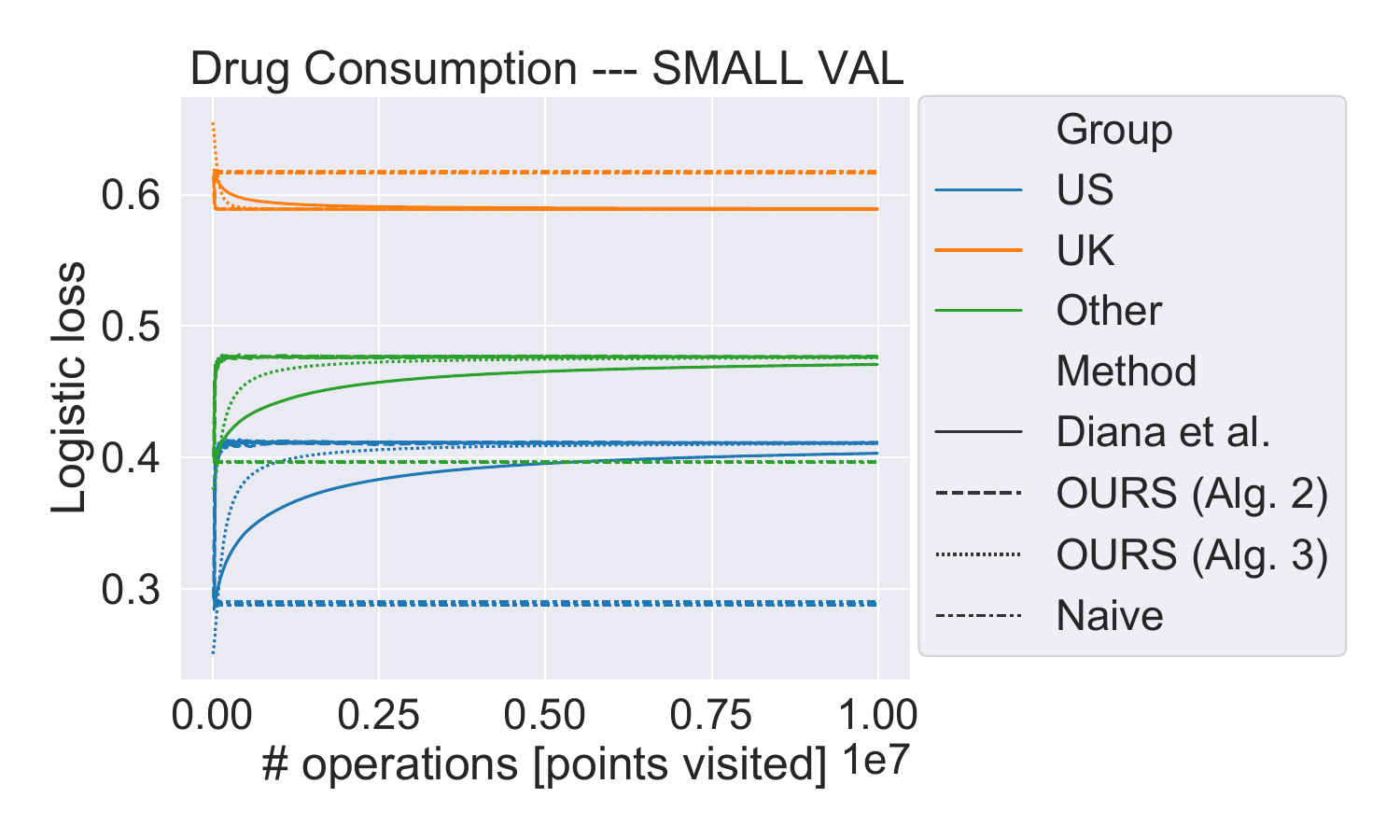}
    \hspace{6mm}
    \includegraphics[height=\heightA]{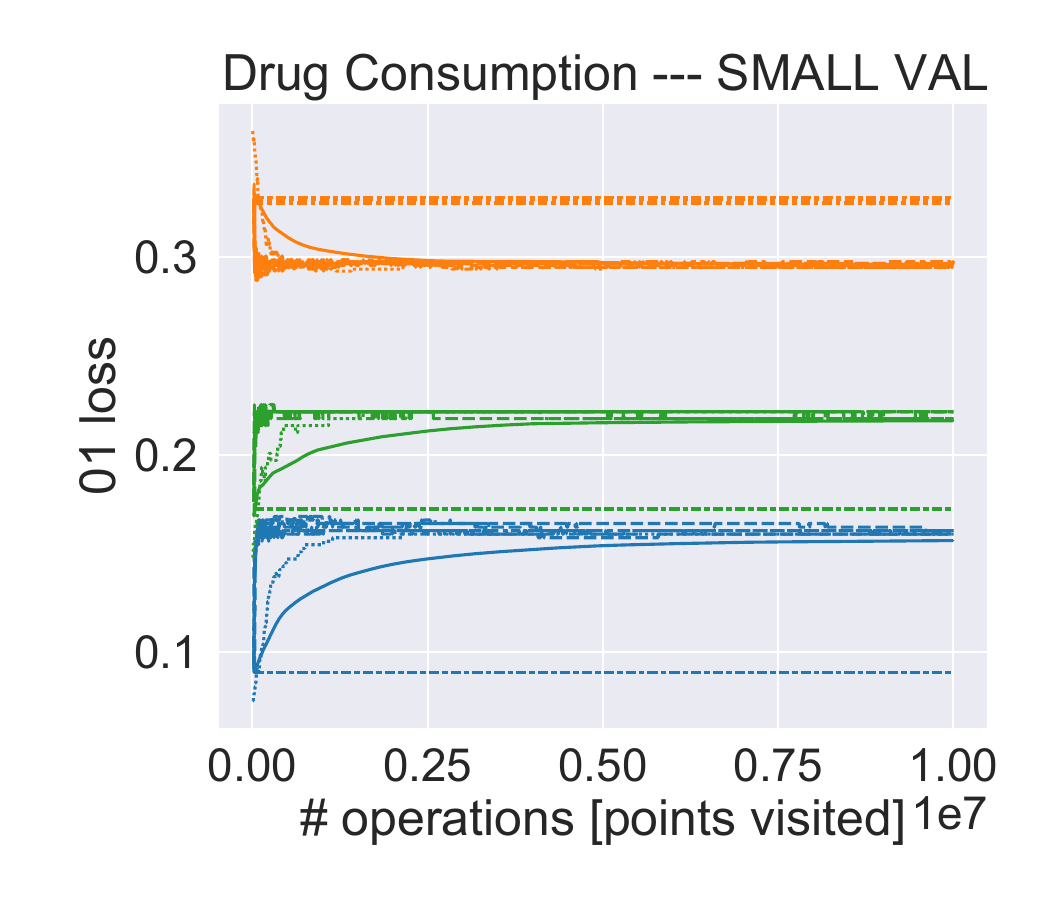}
    \hspace{6mm}
\includegraphics[height=\heightA]{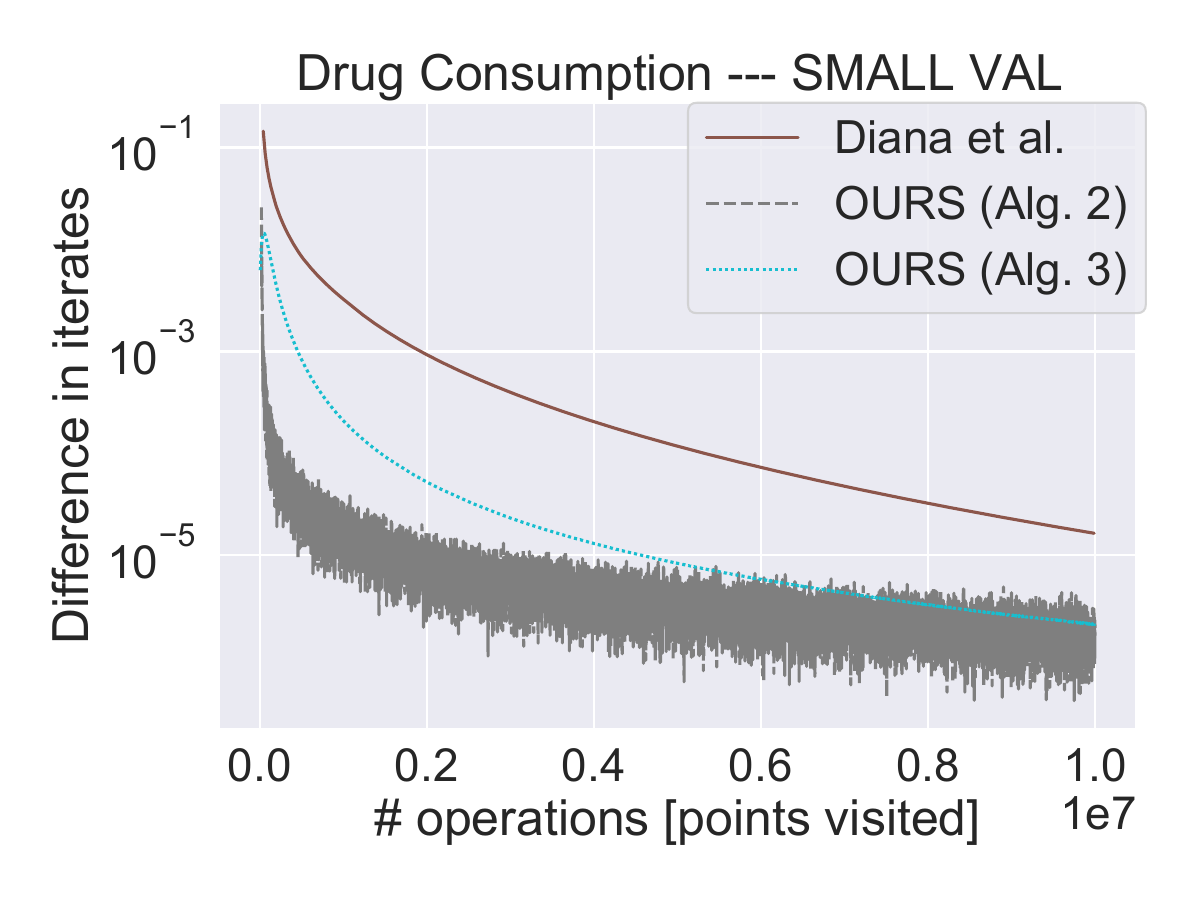}

    \begin{minipage}[t]{3.8cm}
    \vspace{0mm}
    \hspace{-14mm}
     \includegraphics[height=\heightA]{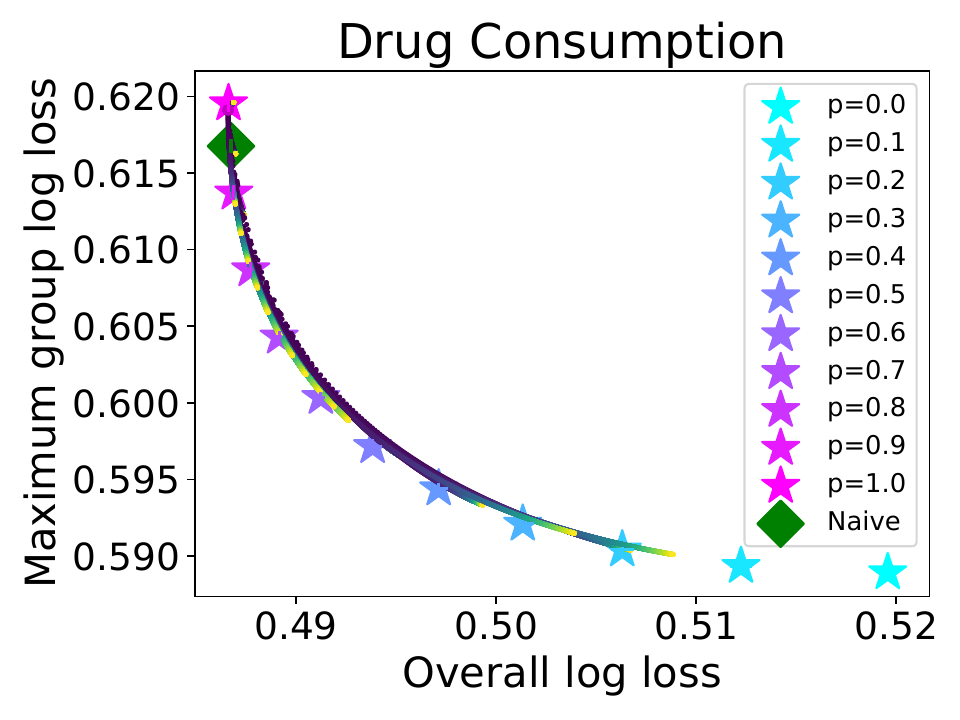}
     \end{minipage}
    \hspace{-2mm}
    \begin{minipage}[t]{9cm}
    \vspace{0mm}
    \begin{scriptsize}
    \begin{tabular}{p{5.3cm}ccccc}
            \toprule
            {} & \makecell{US} & \makecell{UK} & \makecell{Other} & \makecell{Overall} &\\
            \midrule 
            \citet{diana2021} & 0.4035 & \textbf{0.5894} & 0.4710 & 0.5166 &\multirow{3}{*}[-0.2cm]{\rotatebox[origin=c]{270}{\parbox[c]{1cm}{\centering \textbf{Log loss} }}}\\
            \citet{martinez2020} & 0.4122 &  \textbf{0.5889} & 0.4771 & 0.5198 &\\
            OURS (Alg. 2 with TRAIN=VAL; Avg. over 5 runs) & 0.4114 & \textbf{0.5889} & 0.4766 & 0.5195 &\\
            OURS (Alg. 2 with SMALL VAL; Avg. over 5 runs) & 0.4112 & \textbf{0.5889} & 0.4766 & 0.5195 &\\
            OURS (Algorithm~\ref{minimax_sgd-optim}) & 0.4108 & \textbf{0.5889} & 0.4761 & 0.5193 &\\
            \midrule 
            \citet{diana2021} & 0.1566 & \textbf{0.2964} & 0.2173 & 0.2432 & \multirow{3}{*}[-0.27cm]{\rotatebox[origin=c]{270}{\parbox[c]{0.7cm}{\centering \textbf{01 loss} }}}\\
            \citet{martinez2020} & 0.1598 & \textbf{0.2950} & 0.2183 & 0.2435 &\\
            OURS (Alg. 2 with TRAIN=VAL; Avg. over 5 runs) & 0.1634 & \textbf{0.2971} & 0.2218  & 0.2463 & \\
            OURS (Alg. 2 with SMALL VAL; Avg. over 5 runs) & 0.1601 & \textbf{0.2960}  & 0.2211 & 0.2446&\\
            OURS (Algorithm~\ref{minimax_sgd-optim}) & 0.1598 & \textbf{0.2960}  & 0.2183 & 0.2440 &\\
            
            \bottomrule
        \end{tabular}
    \end{scriptsize}
    \end{minipage}

    \caption{Logistic regression on the 
    Drug Consumption 
    dataset. %
    \textbf{Top row:} 
Logistic loss and 
classification error 
    for the three groups over time, both for \citet{diana2021}  and our Algorithms~\ref{minimax_sgd} and~\ref{minimax_sgd-optim}. 
    \textbf{Middle row:} Same as top row,  but for a validation set that comprises only 60 datapoints (20 datapoints sampled uniformly at random from each group). 
    \textbf{Bottom row:} Trade-off curves obtained by varying~$\gamma$ in a variant of the algorithm by \citet{diana2021} and a probability parameter~$p$ with which we sample from the whole population in 
    our 
    Algorithm~\ref{minimax_sgd}. 
    (Average) Per-group 
    losses and errors 
    as well as overall losses and errors 
    from the final iteration are shown in the table. For every method, the 
    maximum 
    loss / error among the groups is shown in bold.}\label{fig:exp_drug_data_Cannabis_Country}
\end{figure*}

\paragraph{Fair ML.}
There is 
a large body of work on fairness
in machine learning \citep{barocas-hardt-narayanan}, 
much of it focusing on supervised learning. Many %
fairness notions 
balance performance measures across different groups 
 \citep[e.g.,][]{hardt2016equality}. 
 These notions 
suffer from the ``leveling-down'' discussed in the introduction. 
Min-max fairness notions have been %
proposed as~a~remedy.

\paragraph{Min-max fairness.
}

\citet{martinez2020} consider the search for  min-max Pareto optimal classifiers and present structural results regarding %
the case of unbounded hypothesis sets. By appropriately reparameterizing the space, they show that one can, in principle, model the case of learning min-max Pareto optimal classifiers over the class of deep neural networks. 
\citeauthor{martinez2020} 
propose an algorithm to find 
optimal classifiers (based on sub-gradient descent), but unlike our work, their proposed algorithm %
has no performance  guarantees, and is not  guaranteed to converge.

\citet{diana2021} propose a multiplicative weights update based method to achieve min-max fairness. While they do not require convexity, they assume access to a weighted empirical risk minimization~(ERM) oracle, and it is unclear how to  implement such oracles in a non-convex setting. Furthermore, the  analysis in \citet{diana2021} is only carried out in the population setting where it is assumed that certain weighted ERM problems can be exactly optimized over the distribution. As a result, their work ignores the complexity of the analysis arising from 
the 
stochastic nature of gradient updates. 
One key contribution of our work is the analysis of gradient-based updates, which allow for more efficient computation and the use of highly-optimised frameworks and tools. %
Finally, at least in the non-convex case, the hypothesis output by~\citet{diana2021} needs to be randomized,  %
which can be
problematic in scenarios strongly affecting 
 people 
 \citep{cotter_neurips2019}. 
 
A previous arxiv version of this paper %
introduced a restricted variant of Algorithm~\ref{alg:basic_outline}, where a single sample was drawn from the worst off group, and each round's model was the global optimum on the current dataset, and analyzed its behaviour.
\citet{shekhar2021} claimed that the sampling scheme 
proposed in our previous draft and closely 
related to our Algorithm~\ref{alg:basic_outline} converges to min-max 
fair 
solutions. While their paper does not impose convexity constraints, their algorithm has no rate of convergence guarantees, and their assumptions (particularly Assumption~2) needed to prove convergence 
frequently fail to hold in practice.\footnote{
Their Assumption~2
states \emph{``For any two distinct attributes~$z, z' \in Z$, we must have $L(z, f^*_z ) < L(z', f^*_z )$, for
any $f^*_z \in \arg \min_{f\in F} L(z, f )$.''}
 This rarely holds in practice. In particular, let $\hat{f}$ be a min-max optimal predictor. Consider a group~$z$ with largest loss under $\hat{f}$: we know that $L(\hat{f}, z) \geq L(\hat{f}, z')$ for any other group $z'$. When this inequality is strict and $L(\hat{f}, z) > \max_{z' \neq z} L(\hat{f}, z')$, we actually know that $L(\hat{f}, z) = L(f^*_z, z)$, or mixing $\hat{f}$ and $f^*_z$ would reduce the min-max risk of $\hat{f}$. So, in many applications of interest (for instance, where the min-max optimal predictor has a unique worst off group), their assumption will not hold---\emph{it never held in any of our experiments.}}
We improve %
those results  empirically and theoretically
and  guarantee fast rates of convergence.%

Min-max fairness has 
also 
been studied in unsupervised settings such as dimensionality reduction 
\citep{samadi2018,tantipongpipat2019} 
and clustering \citep{samira_fair_kmeans} as well as  in federated learning scenarios  \citep{mohri2019agnostic, papadaki2022minimax}.

\paragraph{Min-max optimization.} Many problems beyond fairness can be formulated as min-max optimization problems, and the study of generic methods for solving these remains an active field of research \citep[e.g.,][]{Thekumparampil2019,Razaviyayn2020,Ouyang2021}. 
We are unaware of any generic methods that would be appropriate for our fairness problem with a discrete group variable. %

\paragraph{Group reweighting.}    
Other works study the problem of debiasing a dataset via  group reweighting. \citet{li2019} propose a reweighting scheme to reduce 
representation bias. 
While based on a min-max formulation, their problem setting  is different to ours.
\citet{rolf2021} study structural properties of optimal group allocations for a dataset. They present structural results regarding the nature of optimal allocations,  
but no algorithmic results. 

\citet{agarwal2019} consider a fair regression problem under a bounded group loss constraint, which in their setting is equivalent to finding a min-max fair classifier. Similar to \citet{diana2021}, 
they 
design a near optimal regressor assuming access to a weighted risk minimization oracle that can be optimized exactly on the population. Achieving fairness under bounded loss constraints assuming oracle access has also been studied 
by
\citet{cotter2018two}.

\paragraph{Active sampling.} 
Active / adaptive sampling 
lies at the heart of
active learning \citep{settles_survey}. 
Related to our work is the 
paper 
by \citet{Anahideh2020}. In
each round their sampling strategy queries the label of a datapoint that is both informative and
expected to yield a classifier with small violation of a fairness
measure (they do not consider min-max fairness but mainly 
demographic parity,
which requires 
the classifier's prediction to be independent of group membership). 
Unlike our work, their approach requires training a
classifier for every datapoint which might be queried 
before actually querying a datapoint, 
resulting in a significant computational overhead. Moreover,
their work does not provide any theoretical analysis.
Also related is the paper by \citet{campero2019}, who propose to 
actively collect additional \emph{features} for datapoints to
equalize the performance on different~groups.

\newcommand{\heightAA}{3.4cm}
\renewcommand{\heightA}{3.2cm}
\newcommand{\abst}{3mm}
\begin{figure*}[h]
    \centering
    
    \includegraphics[height=\heightA]{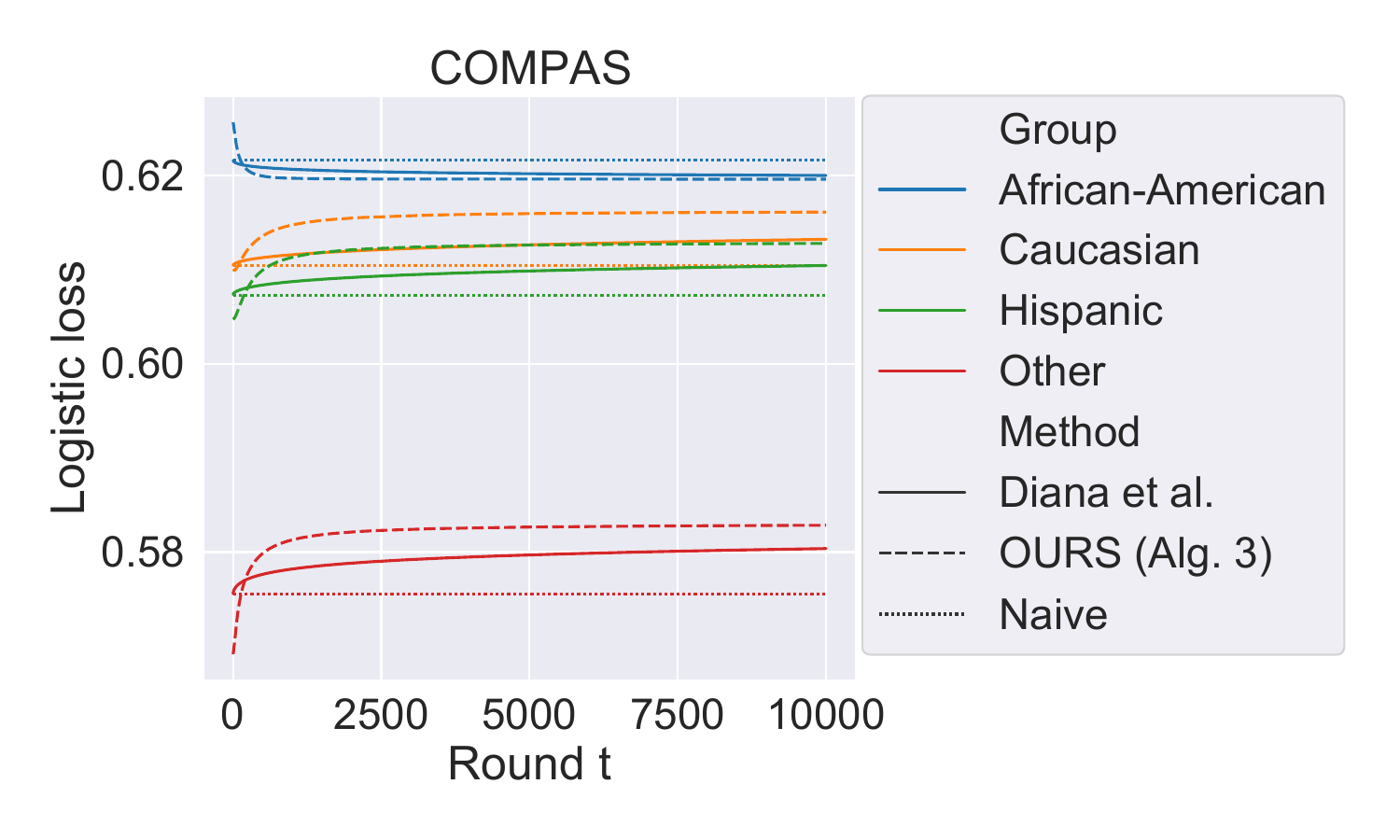}
    \hspace{6mm}
    \includegraphics[height=\heightA]{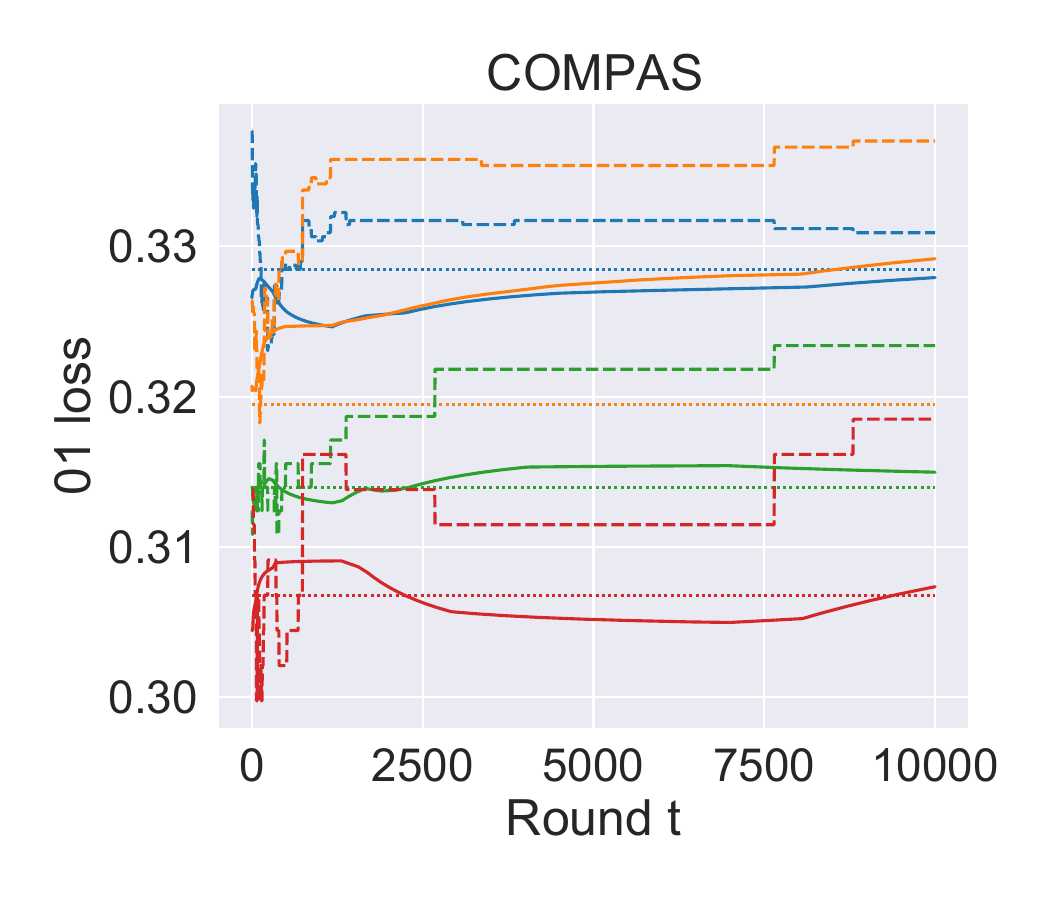}
    \hspace{6mm}
    \includegraphics[height=\heightA]{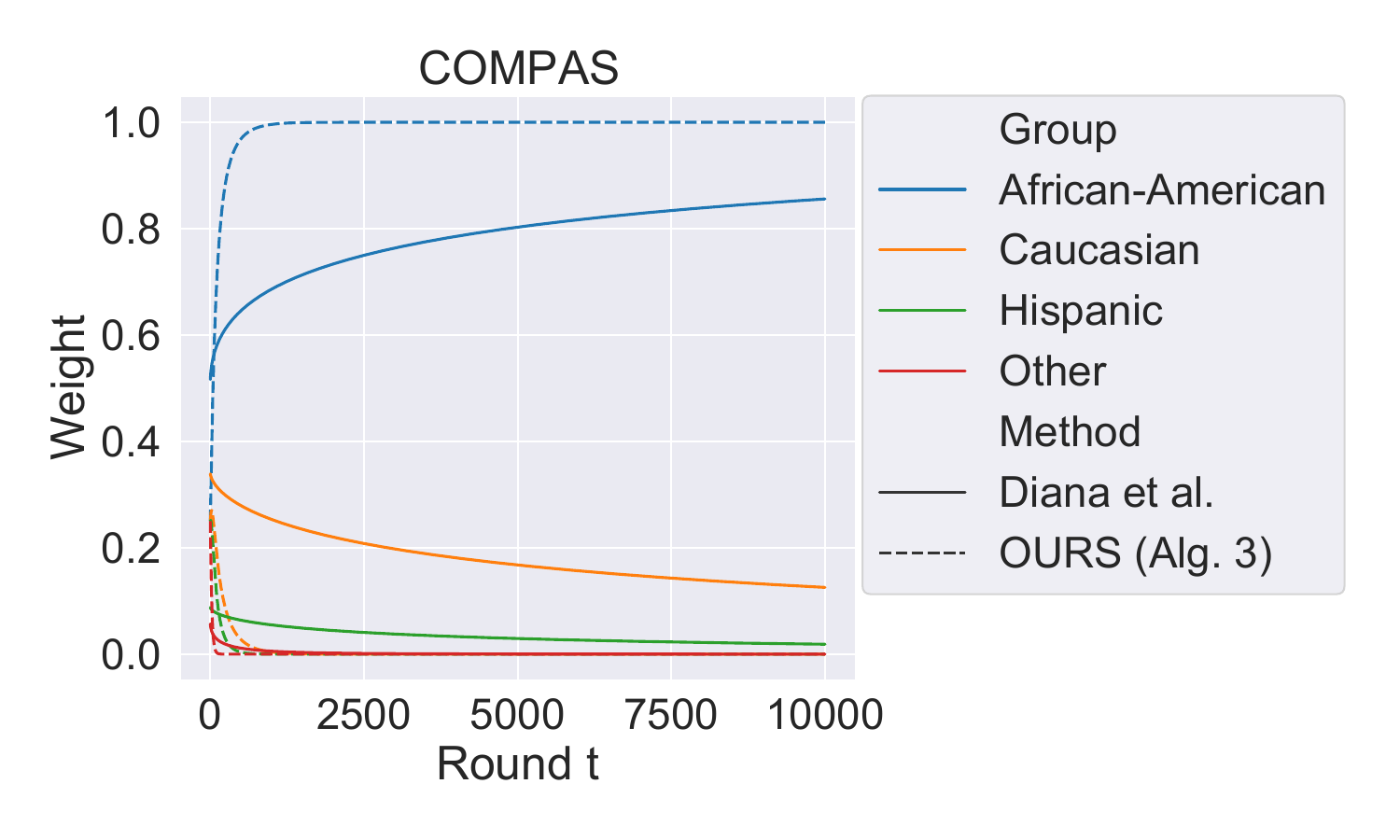}

     \vspace{5mm}
     \begin{scriptsize}
     \renewcommand{\arraystretch}{1.15}
    \begin{tabular}{lcccccc}
           \toprule
            {} & \makecell{African-American}& \makecell{Caucasian}& \makecell{Hispanic}& \makecell{Other}&\makecell{Overall} \\
            \midrule 
            \citet{diana2021} & \textbf{0.6200} & 0.6132 & 0.6104 & 0.5804 & 0.6145 &             \multirow{4}{*}[0.15cm]{\rotatebox[origin=c]{270}{\parbox[c]{0.9cm}{\centering \textbf{Log loss} }}}\\
            \citet{martinez2020} 
            & \textbf{0.6196} & 0.6162& 0.6129 & 0.5830 & 0.6158 &\\
            OURS (Algorithm~\ref{minimax_sgd-optim}) & \textbf{0.6196} & 0.6161 & 0.6127 & 0.5828 & 0.6156 &\\
            \midrule
             \citet{diana2021}  & 0.3279 & \textbf{0.3292} & 0.3150 & 0.3074 & 0.3260 &\multirow{4}{*}[0.12cm]{\rotatebox[origin=c]{270}{\parbox[c]{0.7cm}{\centering \hspace{-2pt}\textbf{01 loss} }}}\\
            \citet{martinez2020} 
            & 0.3309& \textbf{0.3370} &0.3234  & 0.3185  &0.3316 & \\
            OURS (Algorithm~\ref{minimax_sgd-optim}) & 0.3309 & \textbf{0.3370} & 0.3234 & 0.3185 & 0.3316 &\\
            \bottomrule
        \end{tabular}
    \end{scriptsize}

    \caption{
    Logistic regression on the COMPAS dataset. 
     Performance of Algorithm~\ref{minimax_sgd-optim} in comparison to the method by \citet{diana2021}:    
    Logistic loss~(left),  classification error (center), and group weights (right) over time. In the table, for every method we show the 
    maximum 
    loss / error among the groups in bold.}
    \label{fig:exp_COMPAS_appendix}
\end{figure*}

\section{Experiments}\label{section_experiments}

Before presenting empirical results,\footnote{Code available on \url{https://github.com/amazon-research/active-sampling-for-minmax-fairness}}   we 
once more  
highlight 
the key 
advantages of our method over existing ones:
(a) simplicity and computational efficiency---we only
 perform one (stochastic) gradient step in every iteration, while the other methods fully retrain in every iteration; 
(b) stronger convergence guarantees---our proposed Algorithm~\ref{minimax_sgd} (Algorithm~\ref{minimax_sgd-optim}) is guaranteed to converge at a rate of $\sim 1/\sqrt{T}$ ($\sim 1/T$) SGD (GD) steps. In contrast, the algorithm presented in \citet{diana2021} is guaranteed to converge at a rate of $\sim 1/\sqrt{T}$ oracle calls, and \citet{martinez2020} do not prove convergence of their proposed~algorithm.

Our online stochastic approach is substantially faster than  the fully deterministic approaches that exactly solve subproblems at each iteration.
To avoid being mislead by implementation details, such as the choice of implementation language, we consider a proxy for runtime: namely, the number of times any datapoint is examined. Under this metric, the cost of computing a single SGD update of mini-batch size $k$ is $k$; the cost of evaluating the objective w.r.t. a single point is 
$1$; while the cost of evaluating the objective for every datapoint of a dataset of size $n$ is $n$. Logistic regression has a cost of $ni$ where $i$ is the number of iterations needed to reach convergence.

Looking at Algorithm~\ref{minimax_sgd}, we see that a significant bottleneck per iteration is line 3, which evaluates the loss over a validation set. Set size is potentially important: too small and the method may not reliably select the worst off group, but if it is too large, it will needlessly hurt runtime. As such, we evaluate using small validation sets containing $20$ random members of each group, and larger validation sets. For all experiments we use a mini-batch size of 32.

We 
compare 
with the public code of both \citet{diana2021} and \citet{martinez2020}. 
When evaluating efficiency, we focus on the method of \citet{diana2021}, however, since the method of \citet{martinez2020} does not come with theoretical guarantees of convergence (and as such it is incomparable to our methods anyway) and their experimental evaluation does not look at the evolution of iterates over time but only at the final iterate. 
As for our proposed strategy, 
we 
focus on  
Algorithm~\ref{minimax_sgd} as the efficient variant of our 
general 
strategy (Algorithm~\ref{alg:basic_outline}) in the convex case; however, we also study the performance of Algorithm~\ref{minimax_sgd-optim}, and 
we 
run 
Algorithm~\ref{minimax_sgd} \emph{without averaging} %
using a simple neural network as classification model %
to study 
the non-convex case.

Similarly to \citeauthor{diana2021}, we report both 
optimization and 
generalization performance. 
To 
evaluate optimization performance, 
we use small 
datasets and report results on the training data. 
When studying generalization performance, we consider a  large dataset and report results on a 
held-out 
test set. 
Since our strategy 
(Algorithms~\ref{alg:basic_outline} or~\ref{minimax_sgd}) 
is randomized, we show its results for five
runs 
with different random seeds. 
See Appendix~\ref{appendix_detail_implementation} for 
implementation 
details. %

{\bf Heuristics for estimating $W$ and $L$.}
Algorithm~\ref{minimax_sgd-optim} requires  estimates of the values $\theta_1$, $W:= \sup_{\theta \in \Theta} \|\theta - \theta_1\|_2$  and $L := \sup_{\theta \in \Theta}\max_{i \in [g]} \|\nabla_\theta \vloss{\theta}{\grp[i]}\|_2$ in order to set the parameters~$\eta$ and~$\gamma$. We use the same method of estimating them for both experiments shown in Figure~\ref{fig:exp_COMPAS_appendix} and the table of Figure~\ref{fig:exp_drug_data_Cannabis_Country}, respectively: 
as the data is whitened, we simply take $L:=\sqrt{d}$, where $d$ is the number of parameters in our logistic regression model, as an upper bound for the gradient of logistic regression.  For $\theta_1$, we run unweighted logistic regression over the entire dataset and use this as the initialisation of our model. Finally, we take $W=\|\theta_1\|$ as an approximate estimate of the size of the domain.

\begin{table*}
    \centering
    \caption{Logistic regression on the Diabetes dataset.  
 (Average) Per-group 
    log 
    losses and 
    classification 
    errors from the final iteration. For every method, the 
    maximum 
    loss / error among the groups is shown in bold.}
    \label{tab:diabetes}
    \begin{scriptsize}
 \begin{tabular}{p{2.7cm} p{0.7cm}cccccp{1.7mm}|cccccp{0.7mm}}
            \toprule
            {} & {} & \makecell{[0-50)} & \makecell{[50-60)} & \makecell{[60-70)}& \makecell{[70-80)}& \makecell{[80-90)} & {} & \makecell{[0-50)} & \makecell{[50-60)} & \makecell{[60-70)}& \makecell{[70-80)}& \makecell{[80-90)} & {} \\
            \midrule 
            \multirow{2}{*}[0cm]{\citet{diana2021}} & Train & 0.6327 & 0.6425 & 0.6474 & \textbf{0.6550} & 0.6473 &  \multirow{6}{*}[-0cm]{\rotatebox[origin=c]{270}{\parbox[c]{0.9cm}{\centering \textbf{Log loss} }}} & 0.3256 & 0.3418 & 0.3500 & \textbf{0.3626} & 0.3498 &\\ 
            & Test & 0.6357 & 0.6443 & 0.6495 & \textbf{0.6563} & 0.6509 & & 0.3304 & 0.3450 & 0.3537 & \textbf{0.3652} & 0.3556 & \multirow{6}{*}[0.3cm]{\rotatebox[origin=c]{270}{\parbox[c]{0.7cm}{\centering \textbf{01 loss} }}}\\
            \multirow{2}{*}[0cm]{\citet{martinez2020}} & Train & 0.6145 & 0.6219 & 0.6289 & \textbf{0.6458} & 0.6439 & & 0.3228 & 0.3326 & 0.3435 & \textbf{0.3616} & 0.3573 &\\
             & Test & 0.6113 & 0.6263 & 0.6329 & 0.6448 & \textbf{0.6456} & & 0.3198 & 0.3396 & 0.3513 & \textbf{0.3650} & 0.3611 & \\
            \multirow{2}{*}[0cm]{OURS (Alg. 2; Avg. 5 runs)} & Train & 0.6161 & 0.6232 & 0.6299 & \textbf{0.6458} & 0.6439  & & 0.3240 & 0.3333 & 0.3442 & \textbf{0.3618} & 0.3579 &\\
            &Test & 0.6129 & 0.6277 & 0.6337 & 0.6449 & \textbf{0.6455}  & &  0.3197 & 0.3401 & 0.3511 & \textbf{0.3650} & 0.3600 &\\
            \bottomrule
        \end{tabular}
        \end{scriptsize}
\end{table*}

{\bf Performance on smaller datasets.}\label{subsection1_experiments}
 We compare  
Algorithms~\ref{minimax_sgd} and~\ref{minimax_sgd-optim}  
 with %
 \citet{diana2021} and \citet{martinez2020} 
 on the Drug Consumption dataset 
\citep{drug_consumption_data} 
and the
COMPAS dataset \citep{angwin2018}, respectively.
We provide some details about the datasets used in our experiments in Appendix~\ref{appendix_detail_datasets}.

On the 
Drug Consumption 
dataset, 
we 
train a logistic regressor %
to predict  
if an individual  consumed cannabis 
within the last decade or not.
The groups 
that 
we 
want  
to be fair to 
are defined by an individual's 
country. 
The dataset contains 1885 
records. 
We use 
the entire dataset for training (sampling and performing SGD updates) and for reporting performance~metrics. 
We either use the entire dataset 
or a small subset comprising 20 datapoints sampled uniformly at random from each group as validation set (for determining the group with the highest loss).

\renewcommand{\heightA}{3.2cm}
\newcommand{\spaceA}{0mm}
\begin{figure*}[t]
    \centering
    \includegraphics[height=\heightA]{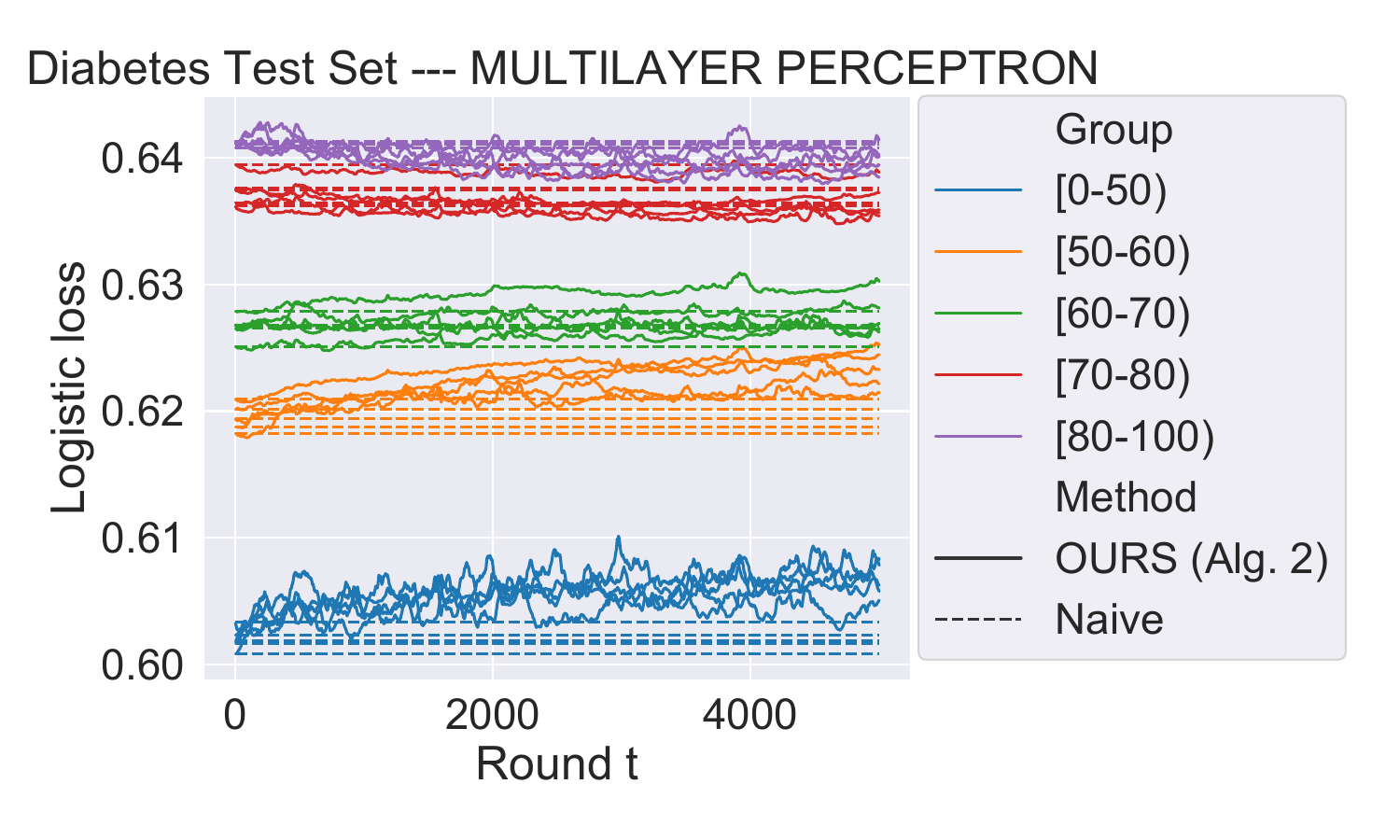}
    \hspace{\spaceA}
   \includegraphics[height=\heightA]{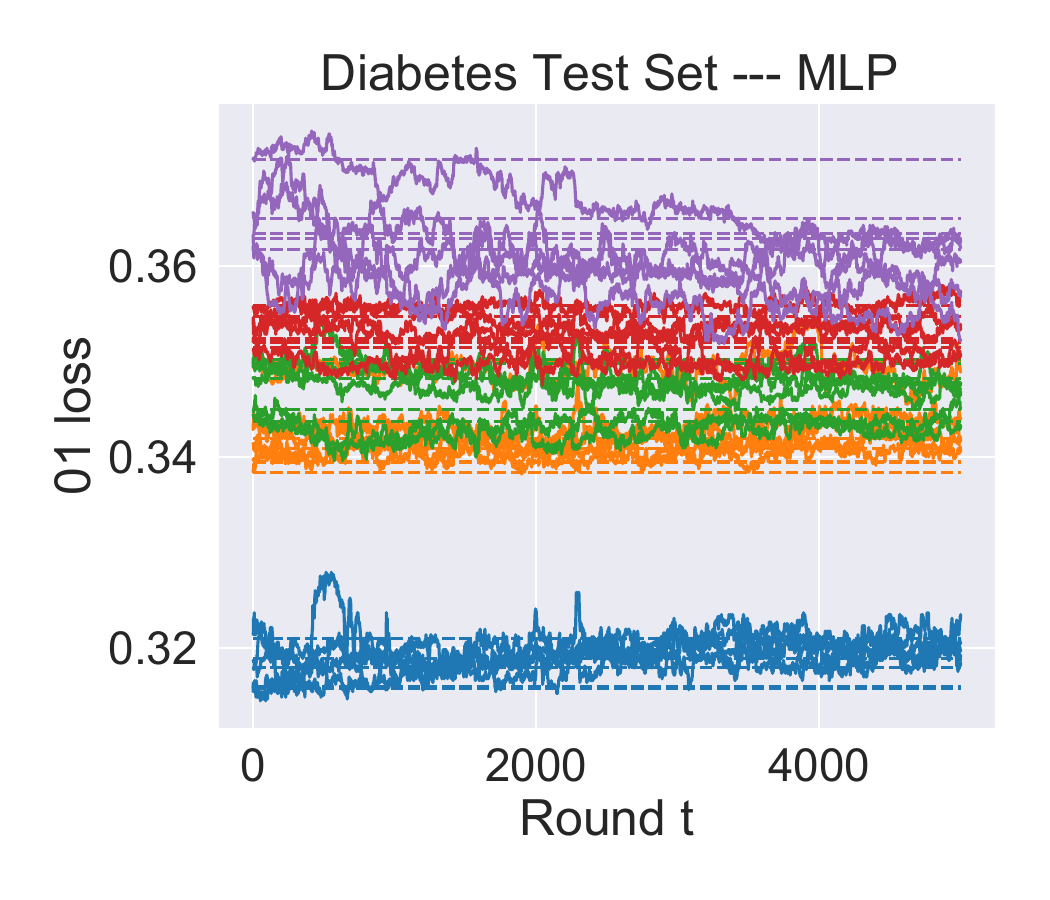}
   \hspace{\spaceA}
   \includegraphics[height=\heightA]{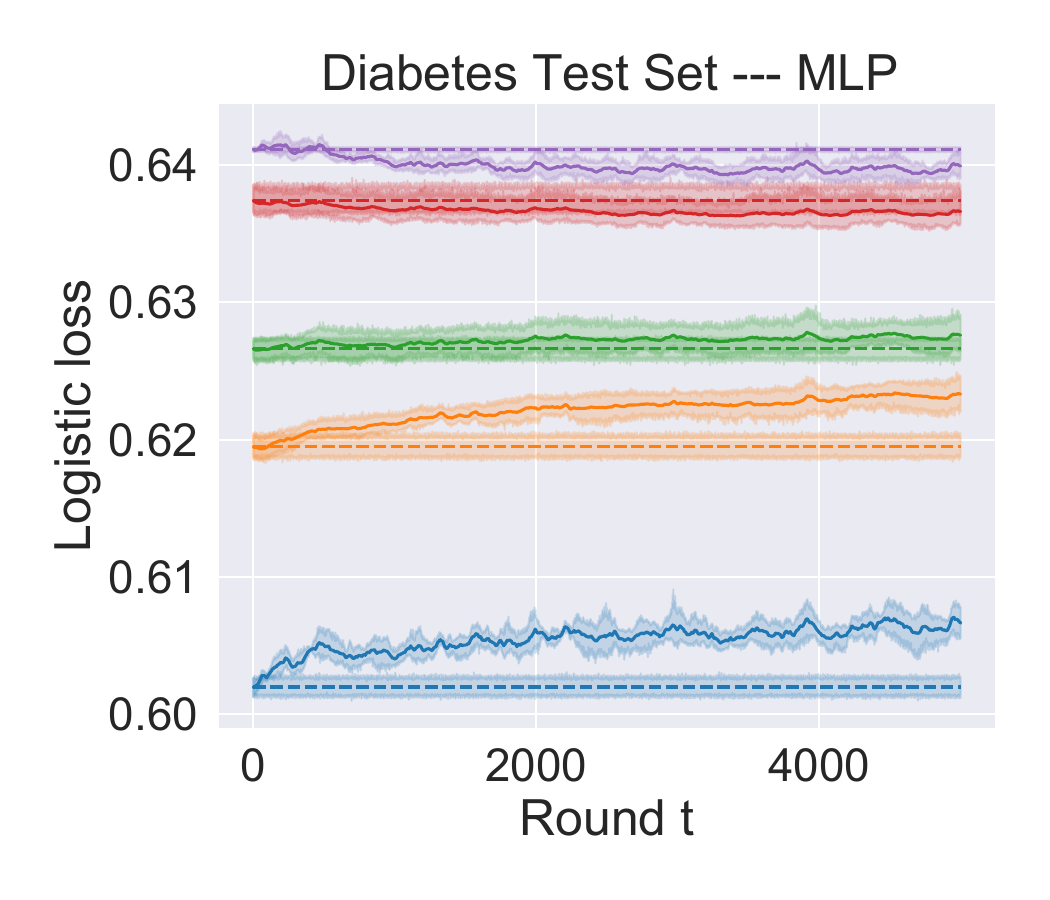}
   \hspace{\spaceA}
   \includegraphics[height=\heightA]{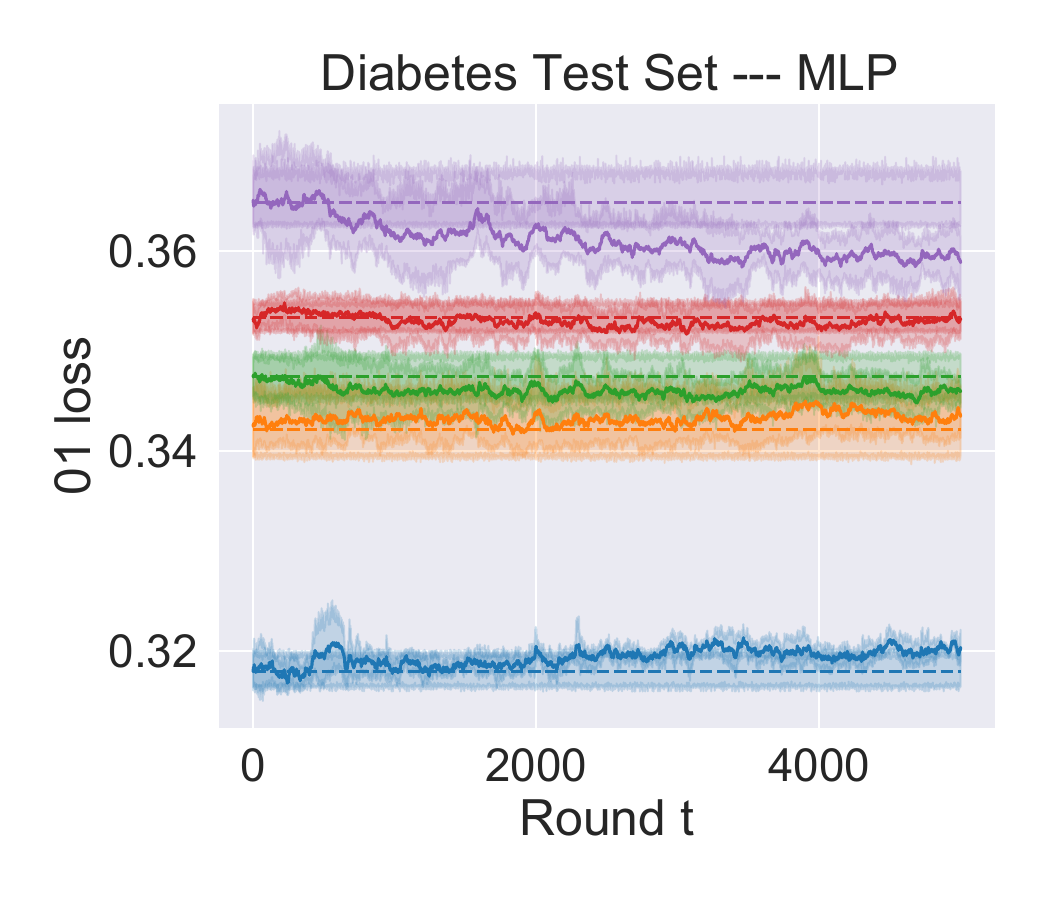}

    \caption{ Algorithm~\ref{minimax_sgd} (returning the final iterate rather than the average) applied to a non-convex MLP 
    on the Diabetes dataset. The plots show the per-group logistic losses and classification errors over time, evaluated on the 
    held-out 
    test set. The first and the second plot show the results for five runs of the experiment; the third and the fourth  show the 
    average results together with 95\%-confidence intervals.}\label{fig:exp_minimax_sgd_optim}
\end{figure*}

Figure~\ref{fig:exp_drug_data_Cannabis_Country} shows the results. %
The plots show~the loss and error for each group over the run of our algorithms or that of \citet{diana2021}. Alongside loss or~error curves, we also plot baseline curves that are obtained by training a classifier or regressor using standard SGD on the training data (dashed lines; denoted by \emph{Naive}).~The~figure also provides a plot showing trade-off curves, trading off the maximum group loss vs the overall 
population 
loss. For \citeauthor{diana2021} the curve is obtained by varying 
the parameter~$\gamma$ in a variant of their algorithm, 
for our strategy we exploit the simple modification discussed in Remark~\ref{remark_mixture_distri}: before determining 
the worst off group, 
we flip a biased coin and with probability~$p$ sample a datapoint from the whole population and 
with probability $1-p$ sample from the worst off group. By varying the parameter~$p\in[0,1]$,~we~generate~the~\mbox{trade-off}~curve. 
The table shows the performance metrics for the model 
 obtained in the final iteration of 
 an algorithm (including the algorithm of \citet{martinez2020}).

The loss (and also the  error) of the worst off group decreases over time, while increasing for other groups. This is consistent with Section~\ref{section_analysis}, which  guarantees improved performance on the highest-loss group, but not for the other groups. In terms of the solution found in the final iteration, we perform similarly to \citet{diana2021} and \citet{martinez2020} with all methods accurately solving the same objective and finding similar cost solutions (cf. the table in Figure~\ref{fig:exp_drug_data_Cannabis_Country}). Also the two trade-off curves are almost identical. 

Very clear trends can be seen in the graphs. While around $1e7$ operations are required for \citet{diana2021} to converge (this is particular apparent in the blue curve representing loss on the US), our approaches converge much faster. In particular, the intrinsic volatility of the SGD update is largely masked by the fast convergence of our approach with all runs converging much faster than any other approach, leading to multiple overlayed plots. The performance benefit is even more extreme when a small validation set is used. Here convergence looks near instantaneous when plotted on a scale that allows us to also see the behavior of the algorithm by \citeauthor{diana2021}. In general, despite its better performance guarantees, our deterministic accelerated version has comparable performance to Algorithm~\ref{minimax_sgd} with a large validation set, and lies midway in performance between Algorithm~\ref{minimax_sgd} with a small validation set and \citet{diana2021}. Note that as the parameters $\gamma$ and $W$ are estimated heuristically, it is likely that better performance could be obtained if they were known; however, we felt it was more informative to report the performance obtained without tuning.

 In  Figure~\ref{fig:exp_COMPAS_appendix}, we 
 evaluate Algorithm~\ref{minimax_sgd-optim} on the COMPAS dataset \citep{angwin2018} and train a logistic regression classifier to predict recidivism. 
 The graphs show a comparison with \citet{diana2021}. 
  For our method a single iteration corresponds to one step of gradient descent, while \citeauthor{diana2021} require the computation of an optimal classifier in each iteration. Despite this, we still converge substantially faster per iteration.

{\bf Generalization  on a larger dataset.}\label{subsection2_experiments}
In Table~\ref{tab:diabetes}, we evaluate on
the Diabetes 130-US Hospitals dataset \citep{diabetes_dataset}. 
The goal is to 
predict 
whether a patient was readmitted to hospital, and we want to be fair with respect to different age groups. We train a linear logistic classifier.  
The 
Diabetes dataset
contains 
101766 records, which we split into a training, validation, and a held-out test set of equal size. The 
latter 
is not used in training.
We initialize our classifier  training on a subset of 2000 training points.

All methods achieve similar loss and error, both when comparing between training and test sets and when comparing the various methods. The former illustrates the   good generalization performance of the algorithms. The results for our Algorithm~\ref{minimax_sgd} and the method by \citet{martinez2020} are even more similar to each other than compared to the method by \citet{diana2021}.

{\bf Algorithm~\ref{minimax_sgd} in non-convex learning.} Minus the averaging step, Algorithm~\ref{minimax_sgd} can also be applied, without guarantees, to non-convex learning problems including neural network training. We
demonstrate
this
by training a simple multilayer perceptron 
(MLP) 
with Algorithm~~\ref{minimax_sgd} on the Diabetes dataset, as in 
the setting of 
Table~\ref{tab:diabetes}. 
    Results are shown in Figure~\ref{fig:exp_minimax_sgd_optim}. The plots show the per-group logistic losses and classification errors on the test set. We improve  the loss and the error of the worst off group 
    compared to 
    the~naive~\mbox{baseline}.

\section{Discussion}\label{sec_discussion}

{\bf Potential harms.} Implicit to  min-max fairness is the idea that the labels are accurate and a trained classifier should reproduce them with high fidelity. As argued by \citet{wachter2021bias}, where this is not the case, for example: where  racially-biased  law  enforcement practices make stop and %
arrest rates a poor surrogate for criminal activity \citep{baumgartner2018suspect}; where hiring data is based on biased historic practices \citep{harvie1998gender}; or when using existing diagnoses to train skin cancer detection \citep{gupta2016skin}; min-max fairness along with other error-based  fairness notions can give rise to classifiers that mimic the biases present in the data, which if used to make substantive decisions about individuals can perpetuate inequality.

{\bf Our contribution.} We  present a novel approach to min-max algorithmic fairness. In contrast to existing approaches, our approach stands out both for its efficient stochastic nature
and easy-to-implement formulations and its guaranteed  convergence rates. 
Our 
experiments on real-world datasets  show the merits of  our approach.

\section*{Acknowledgements}

Jamie Morgenstern and Jie Zhang acknowledge funding from the NSF AI Institute for the Foundations of Machine Learning (IFML), an NSF Career award, and the Simons Collaborative grant on Theory of Algorithmic Fairness.

\bibliography{biblio,mybibfile_fairness,add_bib}
\bibliographystyle{plainnat}

\clearpage
\onecolumn
\normalsize
  
\appendix
\section*{Appendix}

\section{Proof of Theorem~\ref{theorem_advanced_alg}}\label{sec:accel}

Let $\ent{\mixt[]} := -\sum_{i=1}^g \mixt[](i) \log \mixt[](i)$ be the entropy function and $\kl{\mathbf{p}}{\mathbf{q}} := \sum_{i=1}^g \mathbf{p}(i) \log \frac{\mathbf{p}(i)}{\mathbf{q}(i)}$ be the Kullback–Leibler divergence. Let $\mix^\star \in \Delta_g$ be the indicator distribution that  puts all of its mass on $\argmax_{i \in [g]} \vloss{{\thtavg}}{\grp[i]}$. Notice that the iterative description of $\mixt[t]$ allows us to write 
$$\mixt[t] := \argmax_{\mix \in \Delta_g} \frac{1}{\gamma}\ent{\mix} + \sum_{s=1}^t \langle \grplosses[s], \mix \rangle. $$
We first observe that, using Jensen's inequality,
\begin{eqnarray*}
T \max_{i \in [g]} \vloss{{\thtavg}}{\grp[i]} 
    & = & T \vloss{{\thtavg}}{\mix^\star}  \leq   \sum_{t=1}^T \vloss{{\thtt[t]}}{\grpemp[\mix^\star]} 
      =  \left( \sum_{t=1}^T \langle \grplosses[t], \mix^\star \rangle + \frac{\ent{\mix^\star}}{\gamma}\right) - \frac{\ent{\mix^\star}}{\gamma} \\
    & \leq & \left( \sum_{t=1}^T \langle \grplosses[t], \mixt[T] \rangle + \frac{\ent{\mixt[T]}}{\gamma}\right) - \frac{\ent{\mix^\star}}{\gamma} \\
    & = & \langle \grplosses[T], \mixt[T] \rangle + \left( \sum_{t=1}^{T-1} \langle \grplosses[t], \mixt[T] \rangle + \frac{\ent{\mixt[T]}}{\gamma}\right) - \frac{\ent{\mix^\star}}{\gamma} \\
    & = & \langle \grplosses[T], \mixt[T] \rangle + \left( \sum_{t=1}^{T-1} \langle \grplosses[t], \mixt[T-1] \rangle + \frac{\ent{\mixt[T-1]}}{\gamma}\right) - \frac{\kl{\mixt[T+1]}{\mixt[T-1]}}{\gamma} - \frac{\ent{\mix^\star}}{\gamma} \\
    & \cdots & \\
    & = & \sum_{t=1}^{T} \langle \grplosses[t], \mixt[t] \rangle 
    -  \sum_{2=1}^{T} \frac{\kl{\mixt[t]}{\mixt[t-1]}}{\gamma} 
    - \frac{\ent{\mix^\star}}{\gamma} + \frac{\ent{\mixt[1]}}{\gamma} \\
    & \leq &  \sum_{t=1}^{T} \vloss{{\thtt[t]}}{\grpemp[\mixt]} 
    -  \sum_{t=1}^{T} \frac{\|\mixt[t] - \mixt[t-1]\|_1^2}{2\gamma} + \frac{\log g}{\gamma}. 
\end{eqnarray*}
Now let us focus on the first summation on the last line. We notice that the $\theta$ update protocol follows the Optimistic Mirror Descent algorithm \citep{chiang2012online,rakhlin2013optimization}, which leads to the following upper bound that holds for 
arbitrary $\theta_* \in \Theta$:
\begin{equation}
  \sum_{t=1}^{T} \vloss{ \thtt[t] }{\grpemp[\mixt]} 
  \leq \sum_{t=1}^T \vloss{\thtt[*]}{\grpemp[\mixt]} 
  +  \frac{\|\thtt[*] - \thtt[1]\|_2^2}{\eta} 
  + \frac \eta 2 \sum_{t=1}^T\|\nabla \vloss{\theta_t}{\grpemp[\mixt]} 
  - \nabla \vloss{\theta_t}{\grpemp[{\mixt[t-1]}]}\|_2^2.
\end{equation}
We have assumed that $\|\nabla \vloss{\cdot}{\cdot}\|_2$ is uniformly upper bounded by $L$, and therefore it holds that $\|\nabla \vloss{\theta_t}{\grpemp[\mixt]} - \nabla \vloss{\theta_t}{\grpemp[{\mixt[t-1]}]}\|_2 \leq L\|\mixt[t] - \mixt[t-1]\|_1$. We therefore have
\begin{equation}
  \sum_{t=1}^{T} \vloss{{\thtt[t]}}{\grpemp[{\mixt[t]}]} \leq \sum_{t=1}^T \vloss{\thtt[*]}{\grpemp[\mixt]} +  \frac{W^2}{\eta} + \frac {\eta L^2}{ 2} \sum_{t=1}^T\|\mixt[t] - \mixt[t-1]\|_1^2.
\end{equation}
Combining, we have 
\begin{eqnarray*}
\max_{i \in [g]} \vloss{{\thtavg}}{\grp[i]} 
    & \leq &  \frac 1 T \sum_{t=1}^{T} \vloss{{\thtt[*]}}{\grpemp[\mixt]} 
      +  \frac{W^2}{T \eta} + \frac{\log g}{T \gamma} + \frac 1 T \left(\frac{\eta L^2}{2} - \frac 1 {2\gamma} \right) \sum_{t=1}^{T} \|\mixt[t] - \mixt[t-1]\|_1^2.
\end{eqnarray*}
If we set $\gamma = (\eta L^2)^{-1}$, the final summation vanishes. Furthermore, if we let $\bar \mix := \frac 1 T \sum_{t=1}^{T}\mixt[t]$, we see that $$\frac 1 T \sum_{t=1}^{T} \vloss{{\thtt[*]}}{\grpemp[{\mixt[t]}]} =  \vloss{{\thtt[*]}}{\grpemp[{\bar \mix}]} \leq \max_{i\in[g]} \vloss{{\thtt[*]}}{D_i}.$$
Putting it all together gives
\begin{eqnarray*}
\max_{i \in [g]} \vloss{{\thtavg}}{\grp[i]} 
    & \leq &  \max_{i\in[g]} \vloss{{\thtt[*]}}{D_i} +  \frac{W^2}{T \eta} + \frac{\eta L^2 \log g}{T},
\end{eqnarray*}
and plugging in the parameter $\eta = \frac{W}{L\sqrt{\log g}}$ completes the proof.

\section{ Details 
}\label{appendix_partB}

\subsection{Details About Implementation and Hyperparameters}\label{appendix_detail_implementation}

We implemented 
Algorithm~\ref{minimax_sgd}
based on Scikit-learn's (\citealp{scikit-learn}; \url{https://scikit-learn.org}) SGDClassifier 
class,  
and 
we implemented Algorithm~\ref{minimax_sgd-optim} using Pytorch (\url{https://pytorch.org/}). When applying our strategy (Algorithm~\ref{alg:basic_outline}) to the  MLP on the Diabetes dataset, we used Scikit-learn's MLPClassifier class. In that experiment, we used a MLP with two hidden layers of size 10 and 5, respectively.

For the method by \citet{diana2021} we used Scikit-learn's LogisticRegression class with lbfgs-solver as oracle.

\textbf{Regularization parameter}: In the experiments on the Drug Consumption dataset 
and the COMPAS dataset, neither for our algorithms nor for the method by \citet{diana2021}, 
we 
used 
regularization. 
In the experiments on the Diabetes dataset, for our strategy,  we set the regularization parameter for $l_2$-regularization to $10^{-6}$ for the logistic regression classifier and to $10^{-4}$ for the MLP classifier. For the method by \citet{diana2021}, we set the regularization parameter for $l_2$-regularization to $10^{-4}$. When setting it to $10^{-6}$, the learnt classifier does not generalize well to the held-out test set. 
The code of \citet{martinez2020} 
does not provide the option to easily set the regularization parameter via an input argument, and we used their ``hard-coded'' default value of 
$10^{-7}/(2n)$, 
where $n$ is the number of training points, as regularization parameter for $l_2$-regularization.  

\textbf{Learning rate}: In all experiments we used a constant learning rate for our methods. We described how to set the learning rate for Algorithm~\ref{minimax_sgd-optim} in the main body of the paper.  For Algorithm~\ref{minimax_sgd}, we 
used a learning rate of 
0.01 on the Drug Consumption dataset
and 
0.005 (logistic regression) or 0.001 (MLP) on the Diabetes dataset. The codes of \citet{diana2021} or \citet{martinez2020} with logistic regression as the baseline classifier 
do not rely on a learning rate.

We used the same parameters as for our method to train the baseline (naive) classifiers, and in order to perform a fair comparison with our method, we returned the average over the 
iterates instead of the last iterate (by setting the \texttt{average} parameter to True in SGDClassifier; this does not apply to the MLP on the Diabetes dataset). 

In all experiments except on the Diabetes dataset with the logistic regression classifier (cf. Section~\ref{section_experiments}), we initialized our strategy with the baseline classifier. 

All other parameters in the code of \citet{diana2021} or \citet{martinez2020}
are set as their default values.

\subsection{Details About Datasets}\label{appendix_detail_datasets}

In Section~\ref{section_experiments} we use the Drug Consumption dataset \citep{drug_consumption_data}
and the Diabetes 130-US Hospitals dataset \citep[Diabetes dataset;][]{diabetes_dataset}, which are 
both 
publicly available in the UCI repository \citep{uci_repository}. 
We also use the COMPAS dataset \citep{angwin2018}, which is publicly available at
\url{https://github.com/propublica/compas-analysis}.

On the Drug Consumption dataset we use the features \emph{Nscore}, \emph{Escore}, \emph{Oscore}, \emph{Ascore}, \emph{Cscore}, \emph{Impulsive}, and \emph{SS} for predicting whether an individual  consumed cannabis 
within the last decade or not. We define the groups by an individual's country, where we merge Australia, Canada,  New Zealand, Republic of Ireland and Other into one group ``Other''.  
On the Diabetes dataset, we use the features 
\emph{gender}, \emph{age}, \emph{admission\_type\_id},
\emph{time\_in\_hospital}, \emph{num\_lab\_procedures},
\emph{num\_procedures}, \emph{num\_medications}, \emph{number\_outpatient}, \emph{number\_emergency}, \emph{number\_inpatient}, \emph{number\_diagnoses},
\emph{max\_glu\_serum}, \emph{A1Cresult}, \emph{change}, and \emph{diabetesMed} for predicting 
whether a patient was readmitted to hospital or not, and we define the groups by a patient's age.
On the Compas dataset we use the features \emph{age}, \emph{sex}, \emph{priors\_count}, \emph{c\_charge\_degree}, and  \emph{juv\_fel\_count} for predicting 
recidivism. Groups are defined by a person's race, where we merge Asian, Native American and Other into one group ``Other''.

We never provide the group information as a feature.

\end{document}